\documentclass[final]{cvpr}

\usepackage{times}
\usepackage{epsfig}
\usepackage{graphicx}
\usepackage{amsmath}
\usepackage{amssymb}

\usepackage{xcolor}
\usepackage{cite}
\usepackage{amsfonts}
\usepackage{algorithmic}
\usepackage{textcomp}
\usepackage{booktabs}
\usepackage{colortbl}
\usepackage{animate}

\usepackage{tabularx}
\usepackage{xfrac}
\usepackage{hyphenat}
\usepackage{multirow}
\usepackage{nicefrac}

\definecolor{lightgreen}{RGB}{197,224,180}
\definecolor{lightblue}{RGB}{222,235,247}
\definecolor{lightpurple}{RGB}{238,229,241}
\definecolor{vanilla}{rgb}{0.05, 0.05, 0.05}
\definecolor{conv}{rgb}{0.89, 0.26, 0.2}
\definecolor{attention}{rgb}{0.55, 0.66, 0.42}
\definecolor{sqex}{rgb}{0.36, 0.54, 0.68}
\definecolor{exfuse}{rgb}{0.93, 0.57, 0.13}
\definecolor{residual}{rgb}{0.57, 0.33, 0.43}
\definecolor{hybrid}{rgb}{0.62, 0.62, 0.58}
\definecolor{tomato}{rgb}{1.0, 0.44, 0.37}
\definecolor{tuftsblue}{rgb}{0.29, 0.59, 0.82}
\definecolor{yellow-green}{rgb}{0.52, 0.73, 0.4}
\definecolor{airforceblue}{rgb}{0.36, 0.54, 0.66}
\definecolor{atomictangerine}{rgb}{1.0, 0.6, 0.4}
\definecolor{enc}{RGB}{215,143,170}
\definecolor{dec}{RGB}{176,166,217}
\definecolor{enc_h}{RGB}{227,45,145}
\definecolor{dec_h}{RGB}{183,151,207}
\definecolor{bifuse}{rgb}{0.66, 0.69, 0.25}
\definecolor{hohonet}{rgb}{0.66, 0.55, 0.32}
\definecolor{unetpp}{rgb}{0.78, 0.36, 0.75}
\definecolor{attention-unetpp}{rgb}{0.13, 0.57, 0.93}
\definecolor{accessblue}{RGB}{0,105,154}

\newcommand{\first}[1]{\textbf{#1}\cellcolor{lightgreen}}
\newcommand{\second}[1]{#1 \cellcolor{lightblue}}
\newcommand{\third}[1]{#1 \cellcolor{lightpurple}}

\newcommand*{\affaddr}[1]{#1} 
\newcommand*{\affmark}[1][*]{\textsuperscript{#1}}

\usepackage[pagebackref,breaklinks,colorlinks]{hyperref}

\def\confName{ICCV}
\def\confYear{2021}
\graphicspath{{./figs/}{./figs/result/}}

\begin{document}


\title{Hybrid Skip: A Biologically Inspired Skip Connection for the UNet Architecture}

\author{Nikolaos Zioulis\affmark[1,2],
Georgios Albanis\affmark[1],
Petros Drakoulis\affmark[1],\\
Federico Alvarez\affmark[2],
Dimitrios Zarpalas\affmark[1],
Petros Daras\affmark[1]
\\
\affaddr{\affmark[1] Centre for Research and Technology Hellas, Thessaloniki, Greece} \\
\affaddr{\affmark[2] Universidad Polit\'{e}cnica de Madrid, Madrid, Spain}\\
\\
\centering{\small\url{https://vcl3d.github.io/HybridSkip/}}
}

\maketitle

\begin{abstract}
 In this work we introduce a biologically inspired long-range skip connection for the UNet architecture that relies on the perceptual illusion of hybrid images, being images that simultaneously encode two images.
The fusion of early encoder features with deeper decoder ones allows UNet models to produce finer-grained dense predictions.
While proven in segmentation tasks, the network's benefits are down-weighted for dense regression tasks as these long-range skip connections additionally result in texture transfer artifacts.
Specifically for depth estimation, this hurts smoothness and introduces false positive edges which are detrimental to the task due to the depth maps' piece-wise smooth nature.
The proposed HybridSkip connections show improved performance in balancing the trade-off between edge preservation, and the minimization of texture transfer artifacts that hurt smoothness.
This is achieved by the proper and balanced exchange of information that HybridSkip connections offer between the high and low frequency, encoder and decoder features, respectively.
\end{abstract}


\section{Introduction}
\begin{figure}[!htbp]
\captionsetup[subfigure]{position=top,labelformat=empty}

\centering

\subfloat[]{\includegraphics[clip,width=\linewidth]{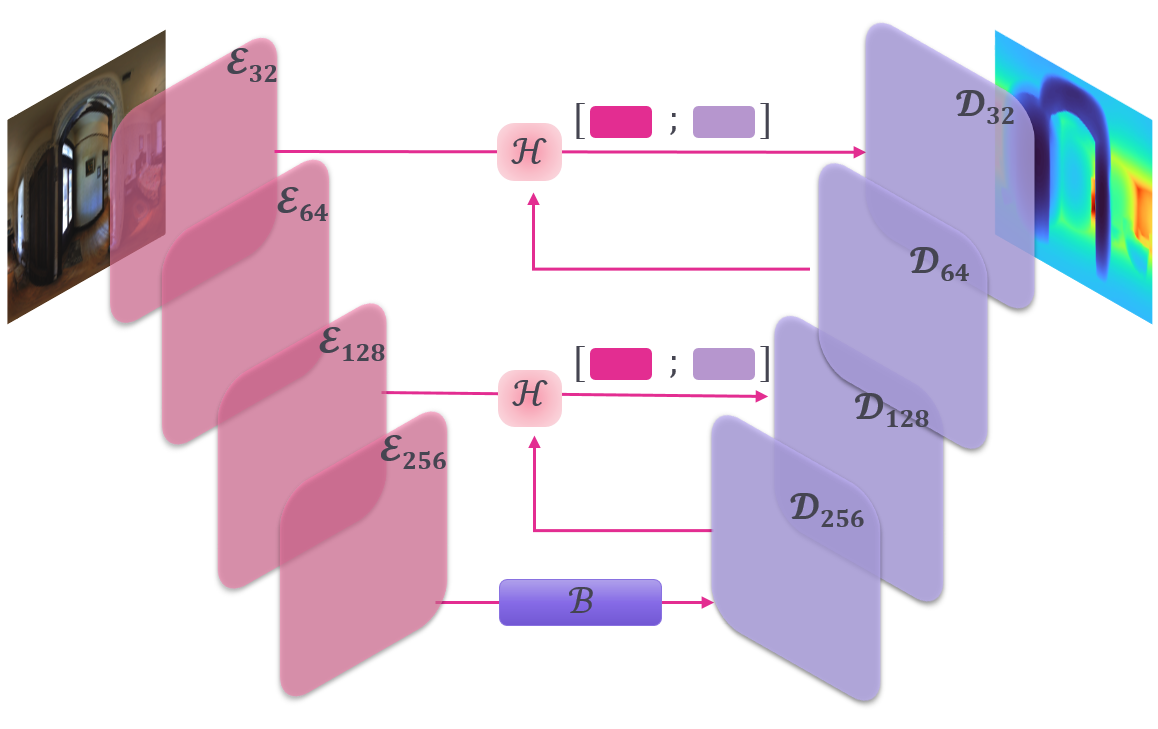}}\\

\vspace{-20pt}

\subfloat[{\color{enc}\rule{\linewidth}{0.5mm}}]{\animategraphics[autoplay,loop,width=0.25\linewidth]{1}{images/enc32/enc_32_Image_}{01}{32}}
\subfloat[{\color{dec}\rule{\linewidth}{0.5mm}}]{\animategraphics[autoplay,loop,width=0.25\linewidth]{1}{images/dec32/dec_32_Image_}{01}{32}}
\subfloat[{\color{enc_h}\rule{\linewidth}{0.5mm}}]{\animategraphics[autoplay,loop,width=0.25\linewidth]{1}{images/enc_h_32/enc_h_32_Image_}{01}{32}}
\subfloat[{\color{dec_h}\rule{\linewidth}{0.5mm}}]{\animategraphics[autoplay,loop,width=0.25\linewidth]{1}{images/dec_h_32/dec_h_32_Image_}{00}{31}}


\caption{
Long-range skip connections are instrumental to the popular UNet architecture but are also challenged by the semantic gap between the \textcolor{enc}{encoder $\mathcal{E}$} and \textcolor{dec}{decoder $\mathcal{D}$} features.
While they allow UNets to capture high resolution details, this is not always beneficial to dense regression tasks that need to overcome texture transfer and also preserve smoothness.
We introduce a biologically inspired skip connection that balances the effect of the high frequency encoder features and the dominant structural information carried by the decoder ones.
From left to right, the bottom row visualizes animations
of the encoder and decoder features maps before and after the hybrid skip connection from a trained dense depth regression model (should they not be playing automatically please consider viewing them with the specific versions of \href{https://get.adobe.com/reader/}{Adobe Acrobar Reader} that support animated images - clicking and holding pauses playback).
}
\label{fig:teaser}
\end{figure}

Skip connections, specifically, the bypassing of convolutional layer blocks within a convolutional neural network (CNN) architecture, are a core building block of modern data-driven models \cite{ye2019understanding}. 
Residual blocks \cite{he2016deep,he2016identity} use short-range skip connections with identity mappings and residual functions to improve information propagation in both forward and backward passes.
They are the basic building block of ResNets, one of the most popular and better performing CNN backends.

At the same time, UNet \cite{ronneberger2015u} is another autoencoder CNN architecture that relies on long-range skip connections, forwarding early encoder features to their corresponding resolution features on the decoder's side.
Different from residual skip connections, UNet concatenates the encoder and decoder features, allowing the network to implicitly learn their fusion through the decoder's convolutional layers.
However, it is a challenging problem as there exists a semantic gap between the encoder features and the corresponding decoder ones, which stems from the higher level concepts and semantic information that is progressively encoded into CNNs. 
Despite this challenge, UNet remains a dominant architecture, especially for semantic segmentation, surpassing fully convolution networks (FCN) \cite{long2015fully}, mainly because it offers higher boundary preservation performance.

Consequently, various works have focused on overcoming this encoder-decoder semantic gap in UNet's long-range skip connections.
Straightforward approaches add learnable operations to lessen the gap with MultiResUNet \cite{ibtehaz2020multiresunet} relying on residual blocks.
More involved approaches utilize gating-based spatial attention \cite{oktay2018attention} to attend to the encoder features in a localized manner, or semantic embedding branches and global convolutions \cite{zhang2018exfuse}.
The search for an appropriate encoder-decoder skip connection led to the use of neural architecture search (NAS) \cite{weng2019unet} to identify the squeeze-and-excite operation \cite{hu2018squeeze} as the more prominent candidate mapping function.

Notably, all these works have applied their proposed skip connection in semantic segmentation, the downstream task that UNet was initially applied at.
Yet recently, UNet-like architectures are increasingly being used in depth estimation as well \cite{mayer2016large,guizilini20203d,ramamonjisoa2021single,godard2019digging,chen2021distortion,zhou2020windowed,9320332}, as the long-range skip connections offer higher boundary preservation performance.
However, the latter is a dense regression task, in contrast to the former, which is a dense classification task.
For depth estimation, the model's parameters encode a continuous function approximation, whereas for semantic segmentation the model focuses on learning a high-dimensional decision surface.
The core difference lies in the nature of depth maps, which are piecewise smooth functions \cite{huang2000statistics}, meaning that compared to semantic segmentation, the smoothness property needs to also hold for the predicted output, whereas for segmentation, the preservation of the boundary is the only secondary trait of importance.
Consequently, for a regression task like depth estimation, skip connections usually result in texture transfer artifacts which hurt smoothness, and introduce false positive boundaries.

In this work, we design a biologically-inspired skip connection based on the way humans process visual input \cite{brady2012spatial,lindeberg2013scale}, specifically the decomposition into different spatial frequencies that happens early on in the visual pathway.
Higher spatial frequencies become imperceptible with farther viewpoints, with the reverse holding for closer viewpoints. The human visual system assimilates higher spatial frequencies into lower ones as viewing distance increases, a mechanism that has been exploited by prior work to generate illusions \cite{oliva2006hybrid}.
We exploit this mechanism as well, to facilitate the exchange of information between the encoder and decoder features, taking into account their higher and lower frequency nature respectively resulting from the autoencoder's inductive bias.

More specifically, we contribute the following:
\begin{itemize}
    \item We design a lightweight and plug-n-play hybrid feature skip connection for the UNet architecture. It performs a blending-based information exchange between the higher and lower level feature maps partaking in a long-range skip connection, prior to their fusion.
    \item We experimentally demonstrate the efficacy of various skip connections in a dense regression task, while taking into account their performance differentials on secondary traits as well; namely boundary preservation and smoothness.
    \item We demonstrate that our proposed skip connection strikes a better balance at boosting direct depth, boundary and smoothness performance, compared to other state-of-the-art skip connections.
\end{itemize}

\section{Related Work}
The UNet CNN architecture \cite{ronneberger2015u} was initially introduced for semantic segmentation and was the first architecture to include long-range skip connections, forwarding information from the encoder to the decoder via feature fusion.
The skip connection improves detail preservation by propagating the early encoder features near the prediction features, boosting semantic segmentation accuracy by allowing for thinner structure segmentation, rendering UNet the standard architecture for this task.
More information about the UNet architecture and the importance of the skip connection can be found in various surveys about U-shaped network architectures \cite{LIU2020244,punn2021modality}.

\begin{figure*}[!htbp]
\centering

\subfloat{\includegraphics[width=0.16\linewidth]{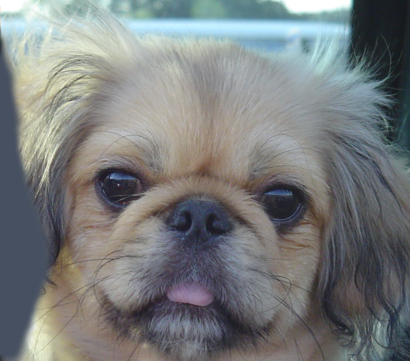}}
\subfloat{\includegraphics[width=0.16\linewidth]{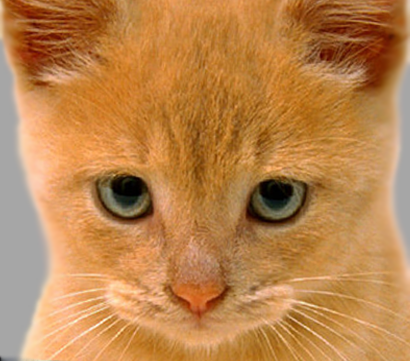}}
\subfloat{\includegraphics[width=0.16\linewidth]{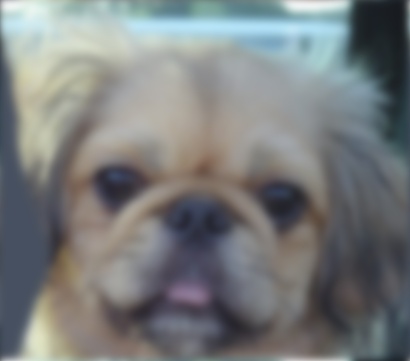}}
\subfloat{\includegraphics[width=0.16\linewidth]{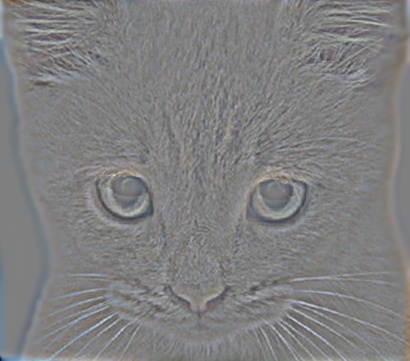}}
\subfloat{\includegraphics[width=0.16\linewidth]{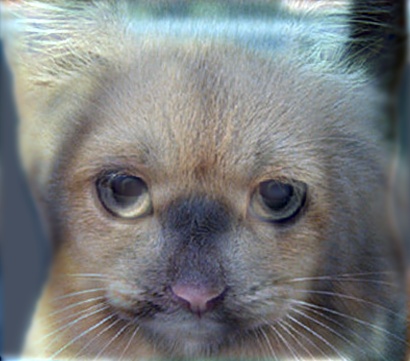}}
\subfloat{\animategraphics[autoplay,palindrome,width=0.16\linewidth]{10}{images/dc/dc_output_}{00}{49}}

\vspace{-10pt}

\subfloat{\includegraphics[width=0.16\linewidth]{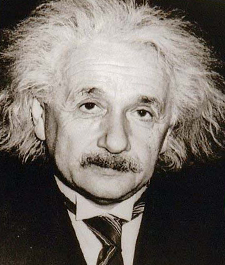}}
\subfloat{\includegraphics[width=0.16\linewidth]{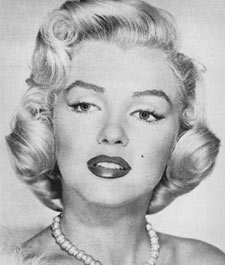}}
\subfloat{\includegraphics[width=0.16\linewidth]{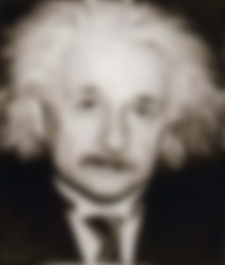}}
\subfloat{\includegraphics[width=0.16\linewidth]{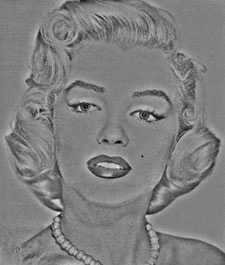}}
\subfloat{\includegraphics[width=0.16\linewidth]{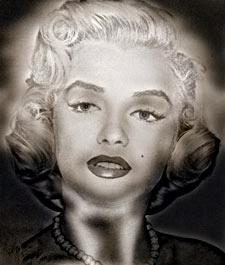}}
\subfloat{\animategraphics[autoplay,palindrome,width=0.16\linewidth]{15}{images/em/em_output_}{00}{49}}

\vspace{-10pt}

\subfloat{\includegraphics[width=0.16\linewidth]{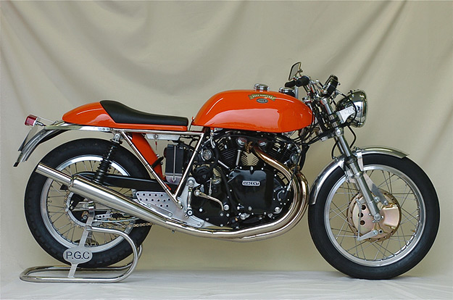}}
\subfloat{\includegraphics[width=0.16\linewidth]{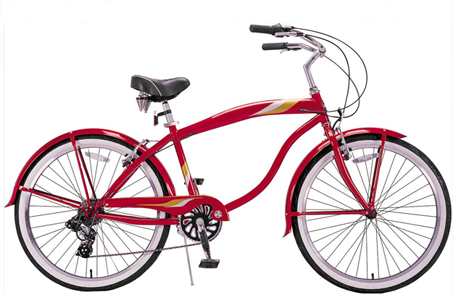}}
\subfloat{\includegraphics[width=0.16\linewidth]{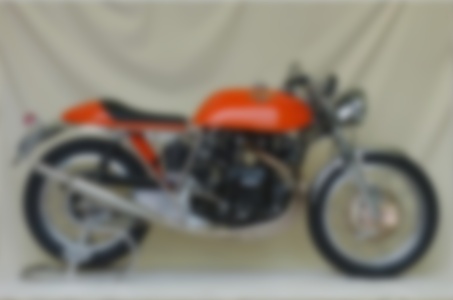}}
\subfloat{\includegraphics[width=0.16\linewidth]{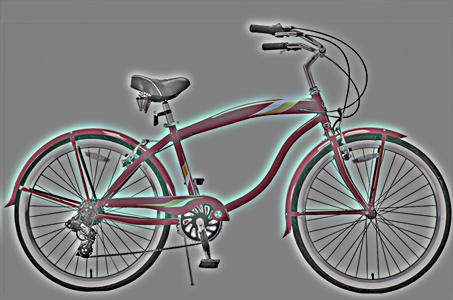}}
\subfloat{\includegraphics[width=0.16\linewidth]{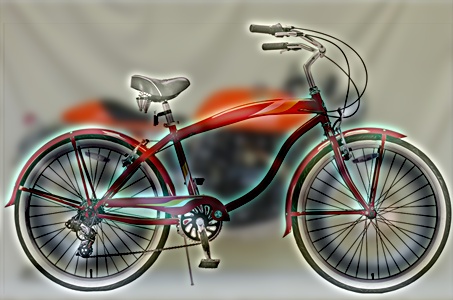}}
\subfloat{\animategraphics[autoplay,palindrome,width=0.16\linewidth]{10}{images/mb/mb_output_}{00}{49}}\\


\caption{
The hybrid images \cite{oliva2006hybrid} human vision based illusion that encode a dual image. 
From left to right: \textbf{i)} first and \textbf{ii)} second image, \textbf{iii)} low pass filtered first image, \textbf{iv)} high pass filtered second image, and \textbf{v)} the hybrid image which changes with viewing distance (\textit{from second to first, by zooming in and out the document respectively}).
The rightmost animated image shows the blending of the low and high pass images using an interpolated blending factor from $0.1$ to $0.9$ that mathematically simulates the physical viewing distance change. (the figure contains animated images, should they not be playing automatically please consider viewing them with the \textit{specific versions of \href{https://get.adobe.com/reader/}{Adobe Acrobar Reader} that support animated images})
}
\label{fig:hybrid_images}
\end{figure*}

Due to its efficacy, it has received a lot of attention and multiple variants have surfaced, with some notable examples being UNet++ \cite{10.1007/978-3-030-00889-5_1}, U\textsuperscript{2}Net \cite{QIN2020107404}, UNet 3+ \cite{9053405}, VNet \cite{milletari2016v}, YNet \cite{10.1007/978-3-030-00934-2_99}, WNet \cite{DBLP:journals/corr/abs-1711-08506} and nnUnet \cite{DBLP:journals/corr/abs-1809-10486}.
Further, it has been gaining traction for tasks other than semantic segmentation such as image reconstruction, with examples being inpainting \cite{Liu_2018_ECCV}, view-synthesis \cite{10.1007/978-3-030-58529-7_37, 10.1007/978-3-030-58452-8_26} and relighting \cite{10.1145/3446328}, as well as depth estimation \cite{guizilini20203d,godard2019digging,zhou2020windowed,9320332}.
Adding to the latter, in the recent Mobile AI 2021 Challenge \cite{Ignatov_2021_CVPR} on single image depth estimation, 7 out of 10 submissions used UNet architectures.
Also, the detail preserving nature of early encoder features allowed its application as the discriminator architecture in high quality synthesis tasks \cite{Schonfeld_2020_CVPR}.

Nonetheless, its strength also presents as one of its main weaknesses.
While skip connections propagate details near the prediction layers, facilitating more detailed dense signal reconstructions, they are not necessarily optimal in their pure identity mapping form.
The reason for this is that the raw fusion of early encoder and late decoder information is hindered by their semantic gap.
CNNs typically extract high spatial-frequency details (\textit{e.g.}~edges, texture, lines) in the early stages, while at the deeper layers the network produces category-specific features representations \cite{badrinarayanan2017segnet,he2016deep}.

Among the techniques designed to address this semantic gap, ExFuse \cite{zhang2018exfuse} used a complex skip connection, replacing the identity mapping with a cascade comprising a semantic embedding branch and a global convolution module.
Results in both an FCN and a UNet demonstrated its efficacy in improving semantic segmentation performance.
Approaching the same problem from another perspective, Attention UNet \cite{oktay2018attention} introduced a novel attention gate as the skip connection.
Each skip connection softly attends to the incoming encoder features using a gating signal.
Initially, additive attention between the projections of the gating signal and encoder features is used to generate an attention grid after aggregating and projecting the result. 
This is then resampled and used to reduce or preserve the importance of the encoder features in a localised manner.
Results in medical segmentation showcased an improvement over vanilla UNet.
The concept was similarly applied to the UNet++ architecture, resulting in Attention UNet++ \cite{li2020attention}, which adapts the gating signal to the nesting levels and shorter skip connections.

More recently, in MultiResUnet \cite{ibtehaz2020multiresunet} the identity mapping skip connection was replaced by a series of residual blocks that aim at alleviating the semantic gap between the encoder and decoder features.
Taking into account that earlier encoder features suffer from a bigger semantic gap, more blocks were used in the earlier features than the ones closer to the bottleneck.
Apart from an improvement in dense and boundary segmentation, the residual skip connections also exhibited robustness to noise.
In a similar fashion, MAPUNet \cite{yang2021mixed}, inspired by UNet++ \cite{10.1007/978-3-030-00889-5_1}, and UNet 3+ \cite{9053405}, exploited multi-scale feature fusion and supervision for monocular depth estimation.
Moreover, a UNet++ variant with residual blocks and dense gated convolution based attention \cite{yu2019free} was used for monocular depth estimation using sparse depth measurements \cite{zhao2021attention}.
Finally, NasUNet \cite{weng2019unet} employed neural architecture search to look for an efficient and effective UNet architecture, a finding shared by \cite{roy2018recalibrating} as well.
Their search resulted in identifying the Squeeze-and-Excite operation \cite{hu2018squeeze} as the most dominant replacement for the standard (identity) skip connection.
Apart from the identity mapping, the search performed included traditional and dilated \cite{7913730} convolutions, as well as separable depthwise convolutions \cite{DBLP:journals/corr/SifreM14}.
Evaluation in different medical segmentation datasets showed performance increases at a fraction of the parameters and reduced memory cost.

Evidently, all aforementioned works focused on segmentation tasks, while all works using UNet's skip connections in reconstruction or regression tasks rely on the vanilla UNet.
Dense regression tasks impose more stringent requirements compared to segmentation tasks, as the predicted signals need to exhibit richer properties.
A notable example are depth images that need to preserve edges and their magnitude, while also varying smoothly in areas where no significant discontinuities manifest \cite{huang2000statistics}.
Compared to previous works, we focus on UNet networks used for regression and holistically assess the efficacy of these advanced skip connections \cite{zhang2018exfuse,oktay2018attention,ibtehaz2020multiresunet,weng2019unet} to improve performance and preserve properties like boundaries and smoothness.
Further, we propose a biological vision inspired skip connection based on scale space theory, that better preserves the output signal's secondary properties simultaneously.

\section{Approach}
Our work focuses solely on the long-range skip connections found in the UNet architecture, and specifically the fusion of features coming from different depths of the model.
Encoder features are learned earlier (shallower) and on higher resolutions, while decoder features are learned later (deeper) and on lower resolutions than the correspondingly encoder ones that they will be fused with.
Our inspiration stems from the Hybrid Images \cite{oliva2006hybrid}. We briefly introduce them in Section~\ref{sec:hybrid_images}, following with our proposed Hybrid Skip connection in Section~\ref{sec:hybrid_skip}.

\subsection{Hybrid Images}
\label{sec:hybrid_images}
Hybrid images $\mathcal{H}$ are dual images that jointly encode two different images, $\mathcal{A}$ and $\mathcal{B}$, but only one is largely perceived. 
Their interpretation changes with viewing distance, creating a smooth optical illusion which has been used to study patients \cite{laprevote2010patients}, face identification \cite{miellet2011local}, create two-layer QR codes \cite{yuan2019two}, or even used for recreational art.
They are generated by the blending of two different spatial resolution images:
\begin{equation}
\label{eq:hybrid_images}
    \mathcal{H} = 0.5 \, \mathbf{f}_{l}(\mathcal{A}) + 0.5 \, \mathbf{f}_{h}(\mathcal{B}),
\end{equation}
where $\mathbf{f}_l$ and $\mathbf{f}_h$ are a low-pass and high-pass filter respectively.
Essentially, image $\mathcal{A}$ is highly blurred, making it visible from farther distances, while image $\mathcal{B}$ is composed by edges, which are only visible from close up.
Figure~\ref{fig:hybrid_images} shows the resulting illusion and intermediate representations.

\subsection{Hybrid Skip Connection}
\label{sec:hybrid_skip}
The UNet architecture's success relies on the long-range skip connection \cite{punn2021modality,ye2019understanding} that fuses early encoder features $\mathcal{E} \in \mathbb{R}^{F \times H \times W}$ with late decoder features $\mathcal{D} \in \mathbb{R}^{F \times H \times W}$.
The typical UNet fusion scheme is a learnable fusion using a convolutional layer receiving as input the concatenation of the encoder and decoder features:
\begin{equation}
    \mathcal{F} = H_i([s(\mathcal{E});\mathcal{D}]),
\end{equation}
where $H(\cdot)$ denotes the convolution function of the $i$th layer, and without loss of generality $s(\cdot)$ denotes the encoder features' skip function, which for the typical UNet is the identity mapping.
It is this multi-scale propagation of earlier encoder features to the late decoder layers that allows UNet architectures to capture finer details.
Yet, there are challenges associated with this fusion scheme, namely the semantic gap between $\mathcal{E}$ and $\mathcal{D}$ as well as the different spatial frequencies of these two feature maps. 

Earlier CNN blocks capture lower level features like lines and edges, while later CNN blocks capture higher level features and concepts, a fact that constitutes their -- straightforward -- fusion an inefficient approach.
Further, earlier encoder features are captured in higher resolutions and contain higher spatial frequencies, while later decoder features contain lower spatial frequencies and are typically upsampled at the skip connection fusion step.
Bilinear interpolation of a lower resolution feature map results in low spatial frequencies \cite{cheng2020high}.

Hybrid images, as represented by Eq.~\eqref{eq:hybrid_images}, blend together two images of different spatial frequencies, toggling the perception of one or the other via how the human visual system's perception changes with viewing distance.
The latter mechanism can be generalized to alpha blending:
\begin{equation}
\label{eq:alpha_blending}
    \mathcal{H}_a(\mathcal{A}, \mathcal{B}) = \alpha \, \mathbf{f}_{a}(\mathcal{A}) + (1 - \alpha) \, \mathbf{f}_{b}(\mathcal{B}),
\end{equation}
where $\mathbf{f}_{a}$ and $\mathbf{f}_{b}$ are two filters converting $\mathcal{A}$ and $\mathcal{B}$ into different frequency images.
The blending coefficient $\alpha$ controls the viewing distance, and therefore, converts the dual image to a distinctly perceived representation.
Figure~\ref{fig:hybrid_images} shows the transition from one image to the other as $\alpha$ is interpolated in $[0.1, 0.9]$.

Considering the skip connection fusing the semantically and spectrally different feature maps $\mathcal{E}$ and $\mathcal{D}$, we rely on the following hybrid feature functions:
\begin{align}
\label{eq:hybrid_features}
    \mathcal{H}_{\boldsymbol{\delta}}^{d}(\mathcal{E}, \mathcal{D}) &= \boldsymbol{\delta} \, \mathcal{D} + (\mathbf{1} - \boldsymbol{\delta}) \, \mathbf{f}_{l}(\mathcal{E}) \\
\label{eq:hybrid_features2}
    \mathcal{H}_{\boldsymbol{\epsilon}}^{e}(\mathcal{E}, \mathcal{D}) &= \boldsymbol{\epsilon} \, \mathcal{E} + (\mathbf{1} - \boldsymbol{\epsilon}) \, \mathbf{f}_{h}(\mathcal{D}),
\end{align}
where $\boldsymbol{\delta}, \, \boldsymbol{\epsilon} \in \mathbb{R}^{F}$ are two alpha blending vectors.
These are combined to form the hybrid skip connection's fusion function:
\begin{equation}
\label{eq:hybrid_skip}
    \mathcal{F}_{hybrid} = H_i([\mathcal{H}_{\boldsymbol{\epsilon}}^{e}(\mathcal{E}, \mathcal{D}); \mathcal{H}_{\boldsymbol{\epsilon}}^{d}(\mathcal{E}, \mathcal{D})]).
\end{equation}
Compared to most other non-identity skip connections \cite{weng2019unet,ibtehaz2020multiresunet,oktay2018attention,zhang2018exfuse}, the hybrid skip connection presented in Eq.~\eqref{eq:hybrid_features}, \eqref{eq:hybrid_features2} and \eqref{eq:hybrid_skip} facilitates a bidirectional information exchange between the encoder $\mathcal{E}$ and decoder $\mathcal{D}$ features, whereas the aforementioned skip connections only focus on bridging the semantic gap between $\mathcal{E}$ and $\mathcal{D}$ by increasing the semantic information carried by the encoder features $\mathcal{E}$.

\textbf{Analyzing HybridSkip.}
There are multiple ways that $\mathcal{F}_{hybrid}$ can be analyzed.
From an attention perspective it can be considered as a mix of heterogeneous feature boosting \cite{wang2017residual} using a soft attention \cite{jetley2018learn} on the respective features.
The decoder features attend to the encoder ones, and vice versa, boosting specific features depending on the blending factors.
While traditional channel attention simply scale entire feature maps (e.g.~the squeeze-and-excite skip connection in \cite{weng2019unet}) and grid based attention only focuses on spatial feature selection (i.e.~\cite{oktay2018attention}), our hybrid approach is distinctly different, albeit it combines these two concepts.
The channel attended encoder(decoder) features boosting the respective channel attended decoder(encoder) are directly related to the spatial information as already learned by the features.

From a spectral processing point of view, it can be considered as a selective alignment or focusing of the spatial frequencies of the blended feature maps.
Considering that the early encoder features $\mathcal{E}$ contain higher frequencies than the upsampled late decoder features $\mathcal{D}$, the second term in Eq.~\eqref{eq:hybrid_features} and \eqref{eq:hybrid_features2} is essentially a band-pass filtered feature map as low/high frequency inputs are passed through a high/low frequency filter.
Therefore, both terms blend inputs from a frequency spectrum lying in the middle of the two opposite end, spatial frequency wise, original feature maps.

Considering the semantic gap, it is apparent that the hybrid skip connection closes the gap in a symmetric fashion by using both inputs to derive the features to be fused.
In contrast to most approaches, it does not seek to close the gap by aligning the encoder features to the decoder ones (e.g.~as in \cite{ibtehaz2020multiresunet,zhang2018exfuse}), but by appropriately blending them.
As $\boldsymbol{\epsilon}$ decreases, the structural edges derived from the decoder features become more dominant in the fused encoder features, accentuating these edges compared to those encoded in $\mathcal{E}$.
Similarly, as $\boldsymbol{\delta}$ decreases, the smoothed detailed edges encoded in the encoder features progressively add texture to the decoder features.
With appropriate blending factors, both directions tend to reduce texture transfer and preserve the edges that matter, leading to a balancing effect between the smoothness and boundary preservation properties of the resulting fused feature maps, and eventually the predicted signal.
Notably, the process is distinct for each feature map, meaning that with $\boldsymbol{\delta}$ and $\boldsymbol{\epsilon}$ being learnable parameters of the model, it encodes a dual representation of these features and learns which one is more appropriate during training.

\begin{figure*}[!htbp]
\centering

\subfloat[$K = 3$]{\includegraphics[width=0.25\linewidth]{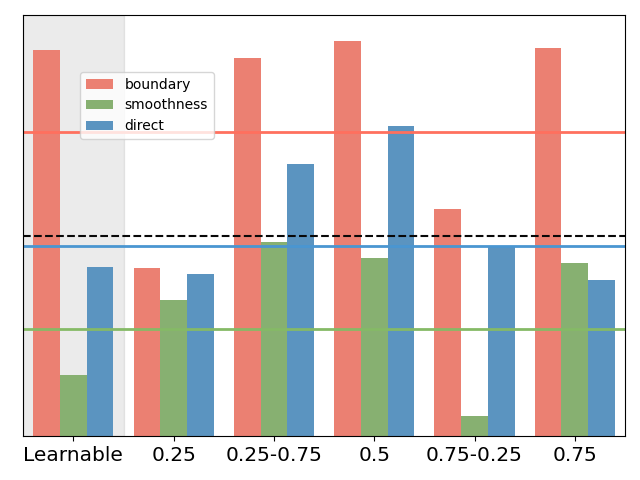}}
\subfloat[$K = 5$]{\includegraphics[width=0.25\linewidth]{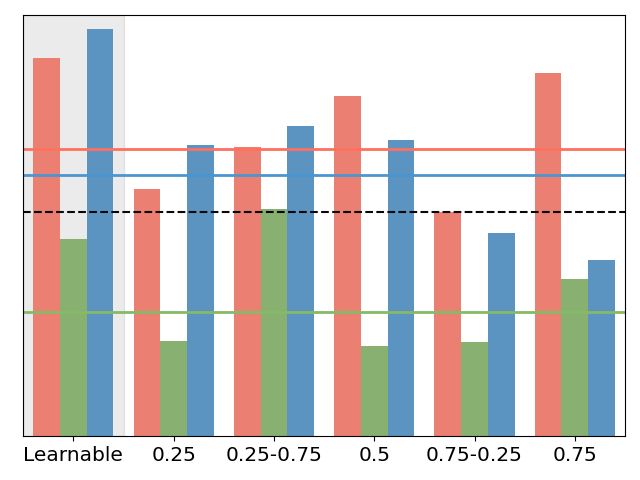}}
\subfloat[$K = 7$]{\includegraphics[width=0.25\linewidth]{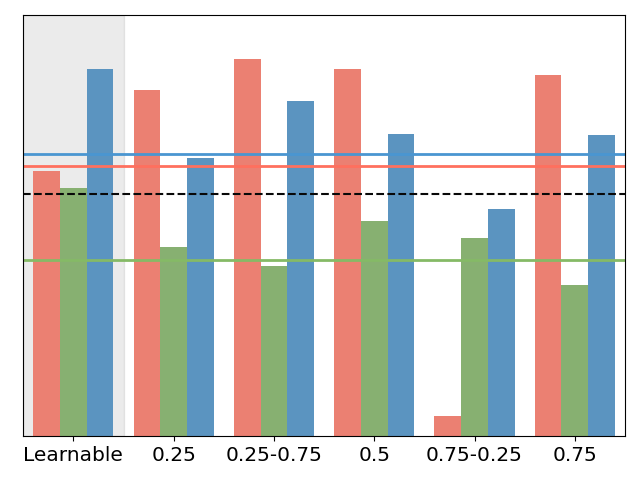}}
\subfloat[$K = 9$]{\includegraphics[width=0.25\linewidth]{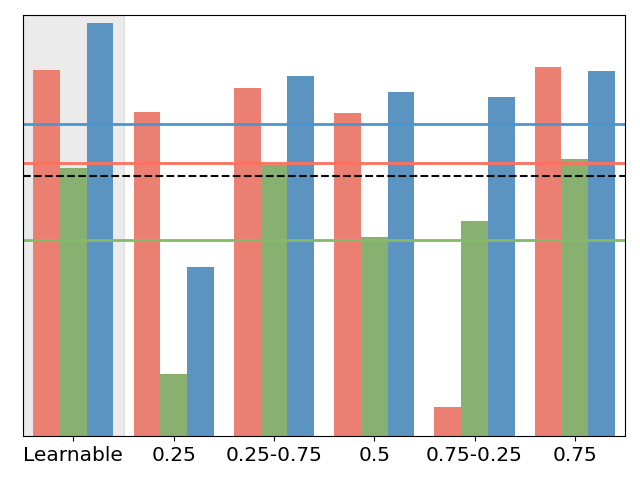}}

\vspace{10pt}

\caption{
Direct depth ($i_d$ - \textcolor{tuftsblue}{pale blue}), boundary ($i_b$ - \textcolor{tomato}{tomato}) and smoothness ($i_s$ - \textcolor{yellow-green}{pale green}) performance indicators across different kernel sizes and blending factors.
The indicator colored horizontal lines denote the average across all blending factors for each kernel size group $K$, while the black dashed line indicates the average of all three performance indicators.
Each bar plots uses a distinct scale which has been normalized to lie in the same range for clarity.
Evidently, with increasing kernel sizes we observe increased average performance, especially for the direct depth and smoothness indicators, while the boundary indicators only slightly benefit from lower kernel sizes.
Nonetheless, when considering all indicators jointly, increasing kernel sizes achieve a balanced and gradual performance increase.
}
\label{fig:ablation}
\end{figure*}
\begin{figure*}[!htbp]
\centering

\subfloat[$F = 512$]{\includegraphics[width=0.2\linewidth, keepaspectratio]{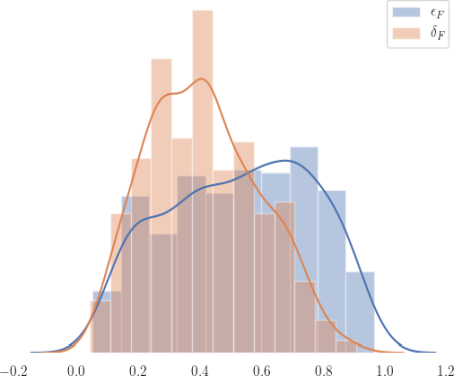}}
\subfloat[$F = 256$]{\includegraphics[width=0.2\linewidth]{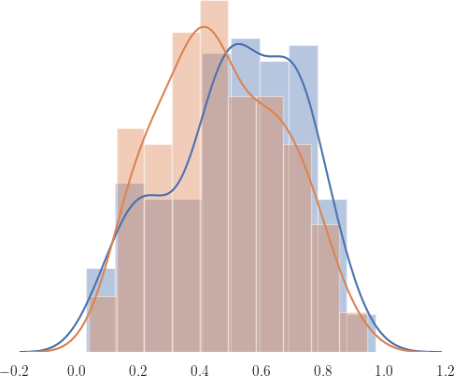}}
\subfloat[$F = 128$]{\includegraphics[width=0.2\linewidth]{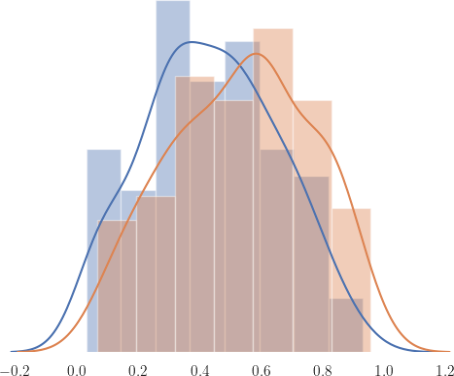}}
\subfloat[$F = 64$]{\includegraphics[width=0.2\linewidth]{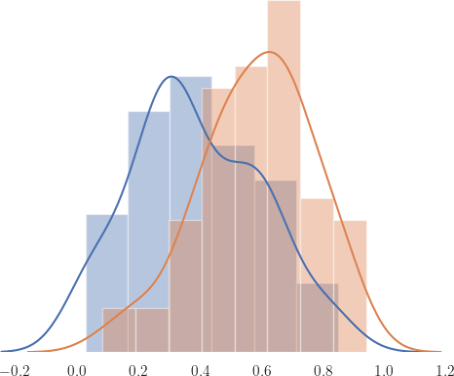}}
\subfloat[$F = 32$]{\includegraphics[width=0.2\linewidth]{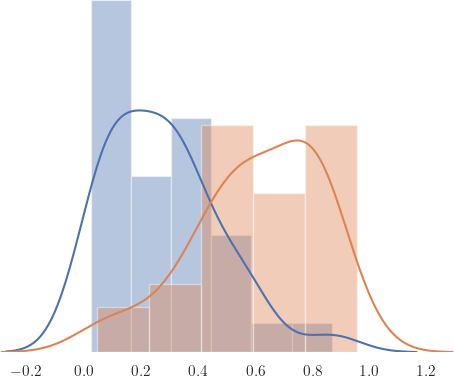}}

\vspace{10pt}

\caption{
The learned \textcolor{airforceblue}{encoder} $\boldsymbol{\epsilon}_F$ and \textcolor{atomictangerine}{decoder} $\boldsymbol{\delta}_F$ blending factors of the $K = 9$ model across the $5$ hybrid skip connections of features $F$. From left to right the model's skip connection transition from the bottleneck to the output layers.
It is observed that as we progress from the bottleneck ($F=512$) towards the prediction layer ($F=32$) a blending factor switch manifests across the HybridSkip connections used in each scale.
The HybridSkip connections closer to the bottleneck focus on the structure given by the encoder (identity and low-pass) features, which nonetheless are closer to the bottleneck, while the skip connections closer to the output focus on the decoder features and their high-pass information.
}
\label{fig:blending_factors}
\end{figure*}

\section{Results}
\begin{figure*}[!htbp]
\centering

\subfloat[\textcolor{conv}{Conv}]{\includegraphics[width=0.3\linewidth]{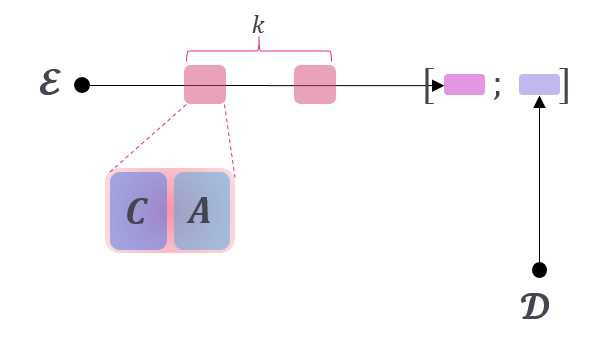}}
\subfloat[\textcolor{residual}{Residual} \cite{ibtehaz2020multiresunet}]{\includegraphics[width=0.3\linewidth]{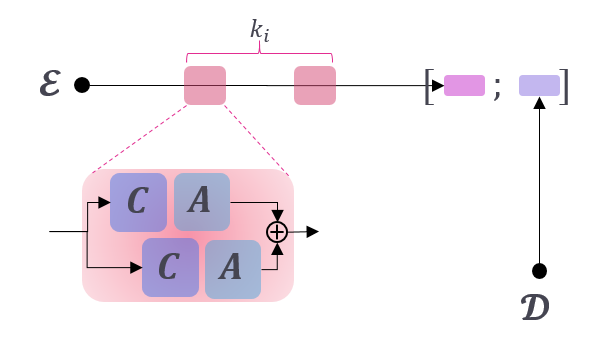}}
\subfloat[\textcolor{exfuse}{ExFuse} \cite{zhang2018exfuse}]{\includegraphics[width=0.3\linewidth]{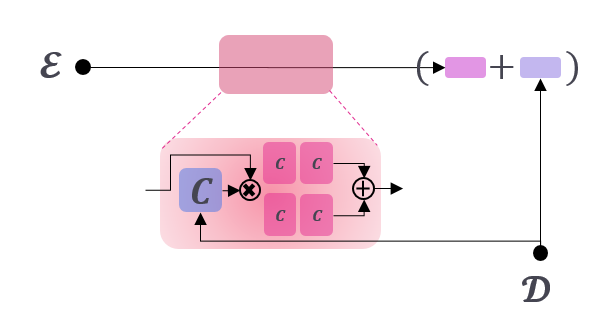}}\\
\subfloat[\textcolor{attention}{Attention} \cite{oktay2018attention}]{\includegraphics[width=0.3\linewidth]{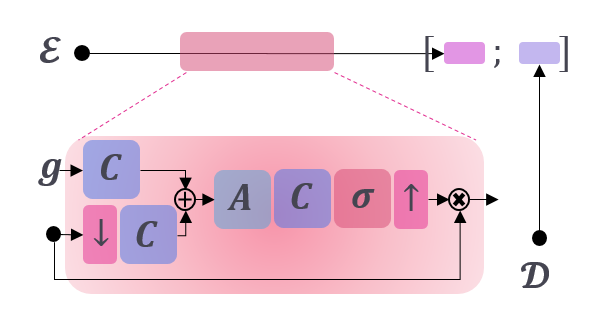}}
\subfloat[\textcolor{sqex}{SqEx} \cite{weng2019unet,hu2018squeeze}]{\includegraphics[width=0.3\linewidth]{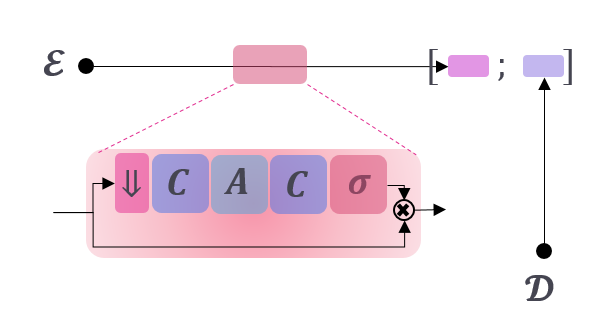}}
\subfloat[\textcolor{hybrid}{Hybrid (Ours)}]{\includegraphics[width=0.3\linewidth]{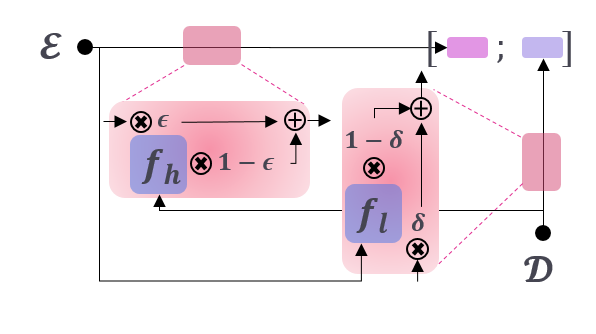}}


\caption{
The architectures of the different skip connections used in the experiments, namely the straightforward convolution layer stack (\textcolor{conv}{Conv}), the residual unit stack (\textcolor{residual}{Residual}), the grid attention skip connection (\textcolor{attention}{Attention}), the Squeeze-and-Excite (\textcolor{sqex}{SqEx}) and \textcolor{exfuse}{ExFuse} skip connections, as well as our proposed \textcolor{hybrid}{Hybrid} one.
The operations included are convolution ($C$), activation ($A$), downsampling ($\downarrow$), upsampling ($\uparrow$), sigmoid ($\sigma$), seperable convolutions ($c$), global average pooling ($\big\Downarrow$), elementwise tensor addition ($\oplus$) and multiplication ($\otimes$), concatenation ($;$), as well as low ($f_l$) and high pass ($f_h$) filtering.
$\mathcal{E}$ and $\mathcal{D}$ denote the input encoder and decoder features of each skip connection, while $\mathit{\mathbf{g}}$ is the gate input used in \cite{oktay2018attention}.
}
\label{fig:skips}
\end{figure*}

\textbf{Experimental Setup.}
For our analysis we use a dense regression task, namely depth estimation, which requires the balancing of both boundary preservation and smoothness of the predicted depth maps, apart from its direct depth estimation performance.
To fully exploit rich depth maps that include both smooth regions, as well as lots of foreground to background depth discontinuities, we use an omnidirectional image benchmark \cite{albanis2021pano3d}.
It includes spherical panoramas that capture entire indoor scenes, containing a lot of flat surfaces (ceiling, floors, tables, etc.), as well as a plurality of foreground objects given their omnidirectional field of view, resulting in a rich piece-wise smooth depth map.
Similar to most works on spherical depth estimation \cite{Pintore_2021_CVPR,eder2019mapped,zioulis2018omnidepth,chen2021distortion,9320332}, we evaluate depths up to $10m$ and use standard metrics for depth estimation, as well as boundary preservation \cite{koch2018evaluation,hu2019revisiting} and surface orientation \cite{wang2020vplnet}.

\textbf{Implementation Details.}
Our implementation is based on \href{https://github.com/ai-in-motion/moai}{\textit{moai}} \cite{moai} which uses PyTorch 1.8 \cite{paszke2019pytorch}, PyTorch Lightning 1.0.7 \cite{falcon2019pytorch} and Kornia 0.4.1 \cite{riba2020kornia}.
For all experiments we use the same UNet architecture and supervision scheme used in Pano3D \cite{albanis2021pano3d}, fixing the learning rate ($0.0002$), optimizer (default parameterized Adam \cite{kingma2014adam}), batch size ($4$) and random number generator seed. 
Thus, only the skip connection varies from experiment to experiment. 
We use the Pano3D low resolution ($512 \times 256$) Matterport3D (M3D) train and test splits for all experiments and apply no data augmentation, training for $60$ epochs.
For the low pass and high pass filters $\mathbf{f}_{l}$ and $\mathbf{f}_{h}$, we use a discrete isotropic Gaussian and a discrete isotropic Laplacian filter respectively.

\subsection{Hybrid Skip Connection Analysis}
\label{subsec:ablation}
In this section we seek to understand the proper design of the hybrid skip connection.
Our analysis focuses on one hand on the kernel size $K$ of the low and high pass filters $\mathbf{f}_{l}$ and $\mathbf{f}_{h}$ respectively, and on the other hand on the choice of the encoder and decoder blending factors $\boldsymbol{\epsilon}$ and $\boldsymbol{\delta}$ respectively.
Regarding the latter, one approach would be to use constant blending factors, explicitly controlling the information exchange between the two feature maps $\mathcal{E}$ and $\mathcal{D}$.

This way, the encoder and decoder blending factors would be $\boldsymbol{\epsilon} = \mathbf{1} * \epsilon$ and $\boldsymbol{\delta} = \mathbf{1} * \delta$, with $\mathbf{1}$ denoting a vector of ones with length $F$ corresponding to the feature maps of each skip connection.
Another approach would be to consider the blending factors as parameters of the model, and jointly optimize them with the convolutional UNet parameters.
This would allow the model to adapt the blending factors to each separate feature instead.
In this case, the blending factors are given by $\boldsymbol{\epsilon} = \sigma(\boldsymbol{\hat{\epsilon}})$ and $\boldsymbol{\delta} = \sigma(\boldsymbol{\hat{\delta}})$, with the hat symbols denoting the model's parameters, and $\sigma$ being the sigmoid function constraining the blending factors to lie in the $[0, 1]$ range.
When using learnable blending factors, the parameters $\boldsymbol{\hat{\epsilon}}$ and $\boldsymbol{\hat{\delta}}$ are initialized using a zero mean and unit variance normal distribution $\mathcal{N}(0, 1)$.

To perform an aggregated analysis among many metrics of different performance traits, namely direct depth, boundary preservation and smoothness, we use the following indicators derived from the metrics used in \cite{albanis2021pano3d}, which aggregate accuracy and error metrics:

\begin{align*}
    i_{d} &= ((1.0 - \delta_{1.25}) \times RMSE)^{-1}\\
    i_{b} &= ((1.0 - \nicefrac{F_1^{1.0} + F_1^{0.25} + F_1^{0.5}}{3}) \times \textit{dbe}^{\text{acc}})^{-1}\\
    i_{s} &= ((1.0 - \nicefrac{\alpha_{11.25^o} + \alpha_{22.5^o} + \alpha_{30^o}}{3}) \times RMSE^o)^{-1},
\end{align*}

\noindent where $F_1^{t}$ are the $F_1$ scores of the precision and recall boundary metrics at each threshold level $t$.
The bar plots in Figure~\ref{fig:ablation} present the results across different kernel sizes and blending factors.
For the former we experiment with $K = \{3, 5, 7, 9 \}$ and for the latter, apart from the learnable blending factors, we also use the following explicit blendings $\{0.25, (0.25, 0.75), 0.5, (0.75, 0.25), 0.75 \}$, with the tuples referring to $(\epsilon, \delta)$ combinations.
Two trends are observed, first, that an increasing kernel size provides consistent performance gains, and second, that the learnable blending factors are also a consistently good performer across different kernel sizes.
Consequently, we use the $K = 9$ kernel size with learnable blending factors as our baseline hybrid skip connection UNet model.

From an interpretation perspective, analysing the learnable blending factors offers an insight on how the hybrid skip connections behave.
We illustrate the distribution of the blending factor coefficients of the $K = 9$ model across its $5$ skip connections in Figure~\ref{fig:blending_factors}.
We observe an interesting and reasonable trend where the deeper layers focus on the structure offered by the incoming encoder features and their low-pass outputs (the encoder features in this case are not early encoder features), while as we progress towards the layers closer to the output, the blending factors indicate that the focus shifts on the predicted signal and its dominant edges, suppressing encoder features resulting into texture transfer.

\begin{table*}[!htbp]
\centering
\caption{Direct depth metrics performance across all compared skip connections. Best three performers are denoted with bold faced
\colorbox{lightgreen}{\textbf{light green}} (1st), \colorbox{lightblue}{light blue} (2nd) and \colorbox{lightpurple}{light purple} (3rd) following the respective ranking order.}
\label{tab:depth_metrics}
\begin{tabular}{l|ccccccccc}
\hline
\multicolumn{1}{c|}{\multirow{3}{*}{UNet Model}} &
  \multicolumn{9}{c}{Direct Depth} \\
\multicolumn{1}{c|}{} &
  \multicolumn{4}{c}{Error $\downarrow$} &
  \multicolumn{5}{c}{Accuracy $\uparrow$} \\
\multicolumn{1}{c|}{} &
  \textit{RMSE} &
  \textit{RMSLE} &
  \textit{AbsRel} &
  \multicolumn{1}{c|}{\textit{SqRel}} &
  $\delta_{1.25}$ &
  $\delta_{1.25^2}$ &
  $\delta_{1.25^3}$ &
  $\delta_{1.05}$ &
  $\delta_{1.1}$ \\ \hline
\textcolor{vanilla}{Vanilla} \cite{ronneberger2015u} &
  0.4055 &
  0.1158 &
  0.1083 &
  \third{0.0649} &
  89.43\% &
  97.34\% &
  99.09\% &
  36.67\% &
  62.12\% \\
\textcolor{conv}{Conv} &
  0.3974 &
  \third{0.0670} &
  0.1095 &
  0.0663 &
  89.43\% &
  97.46\% &
  99.09\% &
  \third{38.53\%} &
  61.79\% \\
\textcolor{attention}{Attention} \cite{oktay2018attention} &
  0.3974 &
  \second{0.0664} &
  \third{0.1074} &
  \second{0.0636} &
  \third{89.67\%} &
  \second{97.61\%} &
  \first{99.19\%} &
  35.90\% &
  61.69\% \\
\textcolor{sqex}{SqEx} \cite{weng2019unet,hu2018squeeze} &
  0.3993 &
  0.0672 &
  0.1097 &
  0.0672 &
  89.57\% &
  97.51\% &
  99.09\% &
  36.11\% &
  61.65\% \\
\textcolor{exfuse}{ExFuse} \cite{zhang2018exfuse} &
  \first{0.3913} &
  0.0865 &
  \second{0.1043} &
  0.0688 &
  \second{90.42\%} &
  \third{97.60\%} &
  99.07\% &
  \first{40.34\%} &
  \first{64.77\%} \\
\textcolor{residual}{Residual} \cite{ibtehaz2020multiresunet} &
  \third{0.3965} &
  0.1068 &
  0.1093 &
  0.0679 &
  89.61\% &
  97.46\% &
  \third{99.13\%} &
  37.58\% &
  \third{62.22\%} \\
\textcolor{hybrid}{Hybrid (Ours)} &
  \second{0.3937} &
  \first{0.0639} &
  \first{0.1010} &
  \first{0.0596} &
  \first{90.76\%} &
  \first{97.72\%} &
  \second{99.17\%} &
  \second{38.75\%} &
  \second{64.41\%} \\ \hline
\end{tabular}%
\end{table*}
\begin{table*}[!htbp]
\centering
\caption{Extra parameters, depth boundary and smoothness preservation metrics. Same colorization scheme as Table~\ref{tab:depth_metrics}.}
\label{tab:secondary_metrics}
\resizebox{\linewidth}{!}{%
\begin{tabular}{lcccccccccc}
\hline
\multicolumn{1}{c}{\multirow{3}{*}{UNet Model}} & \multicolumn{5}{c}{Depth Discontinuity} & \multicolumn{4}{c}{Depth Smoothness} & Model Performance \\ \cline{2-11} 
\multicolumn{1}{c}{} & \multicolumn{2}{c}{Error $\downarrow$} & \multicolumn{3}{c}{Accuracy $\uparrow$} & \multicolumn{3}{c}{Accuracy $\uparrow$} & Error $\downarrow$ & Parameters $\downarrow$ \\
\multicolumn{1}{c}{} & \textit{dbe}\textsuperscript{acc} & \multicolumn{1}{c|}{\textit{dbe}\textsuperscript{comp}} & $F_{1}^{0.25}$ & $F_{1}^{0.5}$ & \multicolumn{1}{c|}{$F_{1}^{1}$} & $\alpha_{11.25^o}$ & $\alpha_{22.5^o}$ & \multicolumn{1}{c|}{$\alpha_{30^o}$} & \multicolumn{1}{c|}{\textit{RMSE\textsuperscript{o}}} & \multicolumn{1}{c|}{} \\ \hline
\multicolumn{1}{l|}{\textcolor{vanilla}{Vanilla} \cite{ronneberger2015u}} &
\second{1.279} &
4.110 &
48.89\% &
42.14\% &
\multicolumn{1}{c|}{32.33\%} &
63.02\% &
77.94\% &
83.12\% &
\multicolumn{1}{c|}{15.95} &
\multicolumn{1}{c|}{27.69M} \\
\multicolumn{1}{l|}{\textcolor{conv}{Conv}} &
\first{1.226} &
4.101 & \third{51.39\%} &
\third{45.92\%} &
\multicolumn{1}{c|}{\third{37.58\%}} &
62.88\% &
78.10\% &
83.35\% &
\multicolumn{1}{c|}{15.83} &
\multicolumn{1}{c|}{+6M (21.66\%)} \\
\multicolumn{1}{l|}{\textcolor{attention}{Attention} \cite{oktay2018attention}} &
1.321 &
\second{3.891} &
51.34\% &
45.66\% &
\multicolumn{1}{c|}{\second{38.05\%}} &
62.60\% &
77.61\% &
82.93\% &
\multicolumn{1}{c|}{15.99} &
\multicolumn{1}{c|}{\third{+2M (8.17\%)}} \\
\multicolumn{1}{l|}{\textcolor{sqex}{SqEx} \cite{weng2019unet,hu2018squeeze}} &
1.344 &
\third{3.931} &
48.83\% &
41.12\% &
\multicolumn{1}{c|}{32.42\%} &
\first{66.22\%} &
\first{79.79\%} &
\first{84.54\%} &
\multicolumn{1}{c|}{\first{14.76}} &
\multicolumn{1}{c|}{\second{+88K (0.31\%)}} \\
\multicolumn{1}{l|}{\textcolor{exfuse}{ExFuse} \cite{zhang2018exfuse}} &
1.528 &
4.865 &
\second{51.60\%} &
\second{46.20\%} &
\multicolumn{1}{c|}{37.20\%} &
\third{63.86\%} &
\third{78.76\%} &
\third{83.84\%} &
\multicolumn{1}{c|}{\third{15.50}} &
\multicolumn{1}{c|}{+18M (64.99\%)} \\
\multicolumn{1}{l|}{\textcolor{residual}{Residual} \cite{ibtehaz2020multiresunet}} &
1.865 &
4.372 &
\first{53.59\%} &
\first{48.28\%} &
\multicolumn{1}{c|}{\first{41.22\%}} &
62.89\% &
78.09\% &
83.38\% &
\multicolumn{1}{c|}{15.71} &
\multicolumn{1}{c|}{+4M (14.44\%)} \\
\multicolumn{1}{l|}{\textcolor{hybrid}{Hybrid (Ours)}} &
\third{1.312} &
\first{3.733} &
49.41\% &
42.94\% &
\multicolumn{1}{c|}{34.42\%} &
\second{64.24\%} &
\second{78.82\%} &
\second{83.86\%} &
\multicolumn{1}{c|}{\second{15.36}} &
\multicolumn{1}{c|}{\first{+1K (0.01\%)}} \\ \hline
\end{tabular}
}
\end{table*}
\begin{figure*}[!htbp]
\centering
\captionsetup[subfigure]{justification=centering}

\subfloat[\textcolor{vanilla}{Vanilla~\cite{ronneberger2015u}\\$(0.218)$}]
{\includegraphics[width=0.165\linewidth]{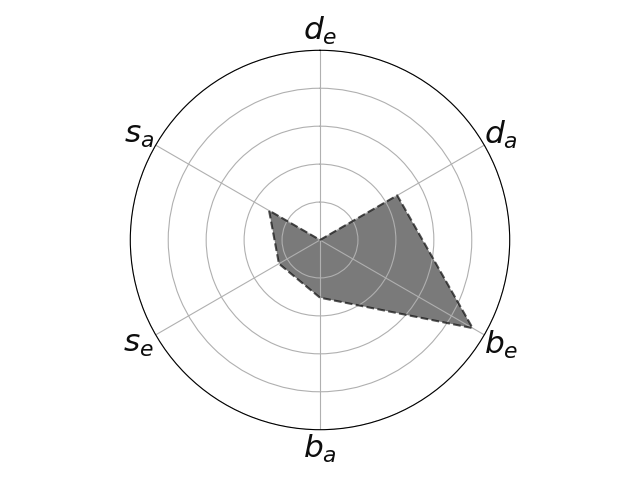}}
\subfloat[\textcolor{conv}{Conv\\$(0.604)$}]
{\includegraphics[width=0.165\linewidth]{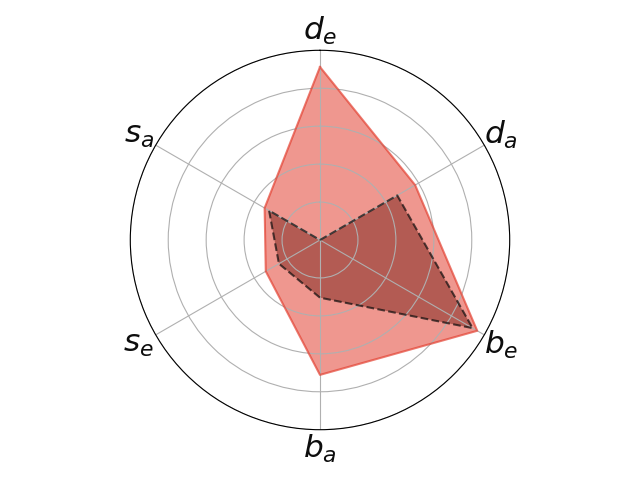}}
\subfloat[\textcolor{attention}{Attention~\cite{oktay2018attention}\\$(0.471)$}]
{\includegraphics[width=0.165\linewidth]{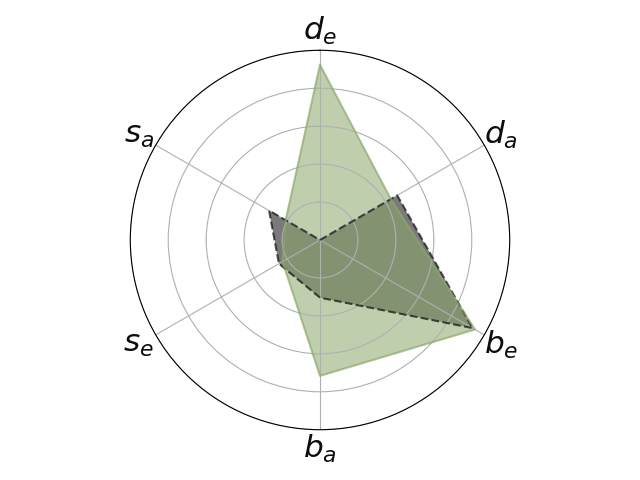}}
\subfloat[\textcolor{sqex}{SqEx~\cite{hu2018squeeze,weng2019unet}\\$(0.798)$}]
{\includegraphics[width=0.165\linewidth]{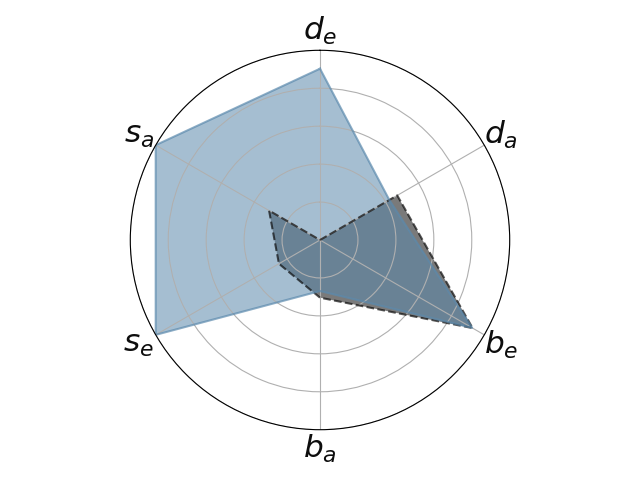}}
\subfloat[\textcolor{exfuse}{ExFuse~\cite{zhang2018exfuse}\\$(0.674)$}]
{\includegraphics[width=0.165\linewidth]{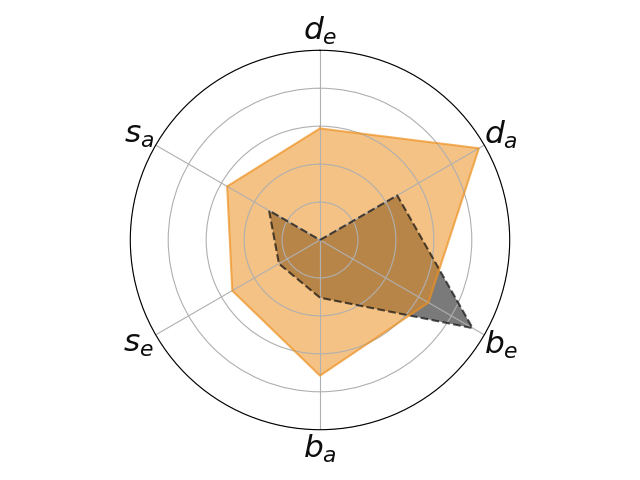}}
\subfloat[\textcolor{residual}{Residual~\cite{ibtehaz2020multiresunet}\\$(0.404)$}]
{\includegraphics[width=0.165\linewidth]{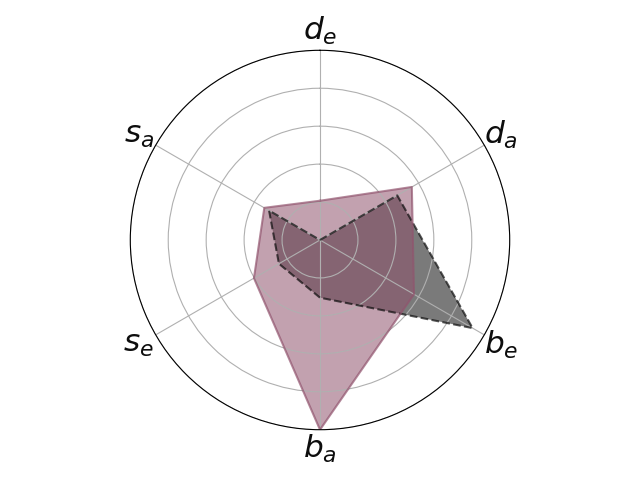}}
\\
\subfloat[\textcolor{hybrid}{Hybrid $(0.842)$}]
{\includegraphics[width=0.165\linewidth]{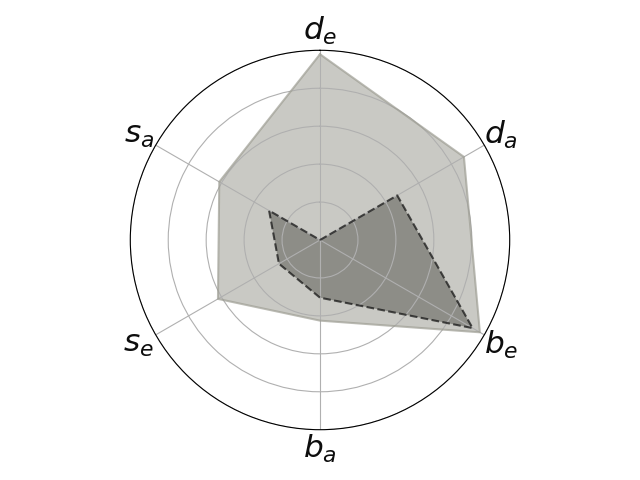}}
\subfloat[\textcolor{hybrid}{Hybrid} vs \textcolor{conv}{Conv}]
{\includegraphics[width=0.165\linewidth]{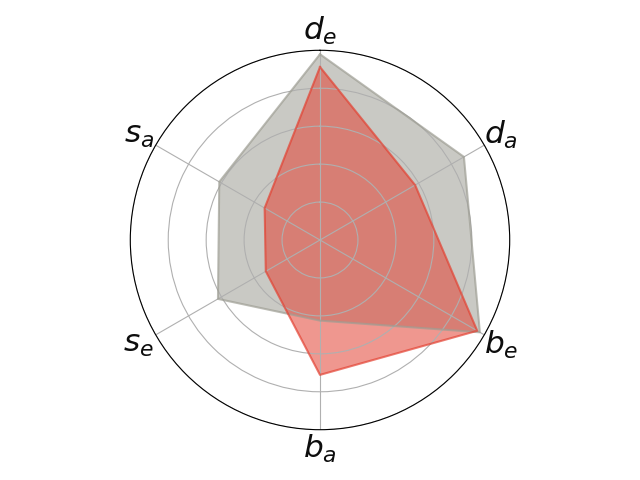}}
\subfloat[\textcolor{hybrid}{Hybrid} vs \textcolor{attention}{Attention}]
{\includegraphics[width=0.165\linewidth]{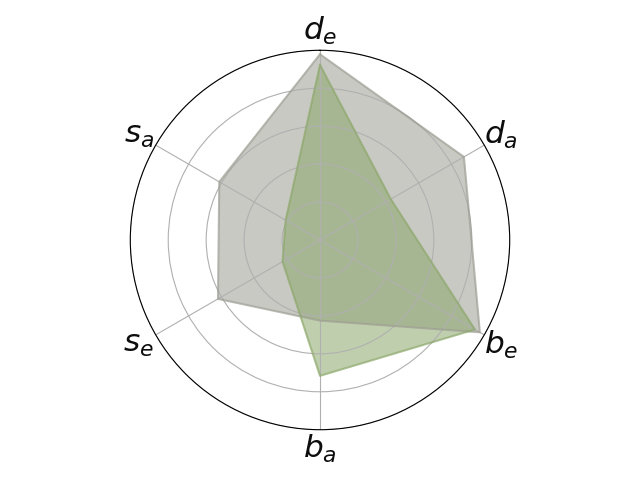}}
\subfloat[\textcolor{hybrid}{Hybrid} vs \textcolor{sqex}{SqEx}]
{\includegraphics[width=0.165\linewidth]{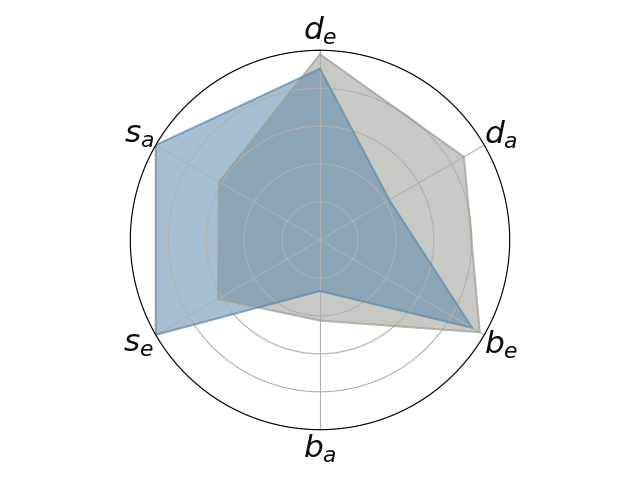}}
\subfloat[\textcolor{hybrid}{Hybrid} vs \textcolor{exfuse}{ExFuse}]
{\includegraphics[width=0.165\linewidth]{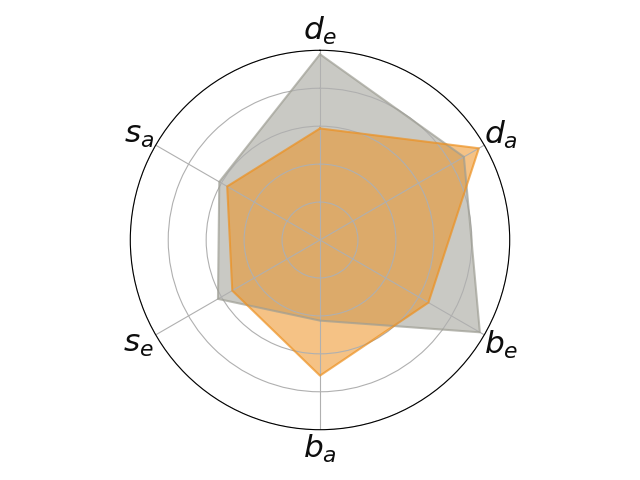}}
\subfloat[\textcolor{hybrid}{Hybrid} vs \textcolor{residual}{Residual}]
{\includegraphics[width=0.165\linewidth]{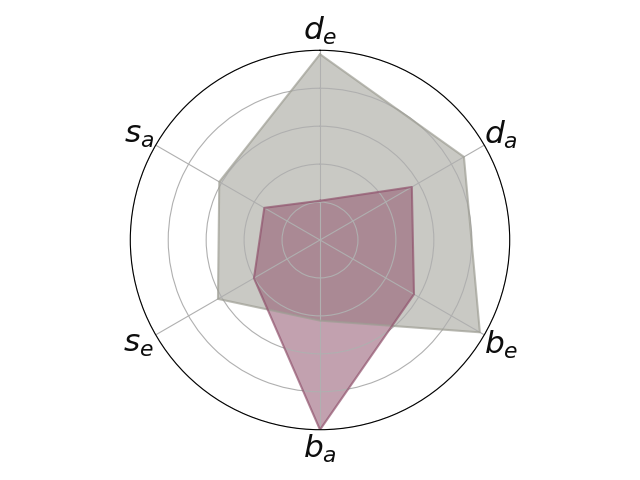}}


\caption{
Normalized performance indicators for direct depth ($d$), boundary ($b$) and smoothness ($s$) preservation accuracy ($a$) and error ($e$) metrics.
The numbers inside the parenthesis indicate the area covered by each different approach, with the largest area of the hybrid skip connection indicating its more balanced performance across all traits.
}
\label{fig:spider}
\end{figure*}

\subsection{Comparison with other Skip Connections}
\label{subsec:comparison}
We additionally compare the performance of the proposed hybrid skip connection to other approaches used for long range skip connections.
More specifically, we present results for a straightforward convolutional (Conv) skip connection stacking $k$ $3 \times 3$ convolution layers, and the stacked residual unit skip connection \cite{ibtehaz2020multiresunet} (Residual), where $k_i$ units are stacked, with $i \in \{1, ..., 5\}$ indicating the $i$th encoder-decoder layers.
Apart from the stacked approaches, we also compare against the attention UNet \cite{oktay2018attention} skip connection (Attention), and the NAS identified \cite{weng2019unet} Squeeze-and-Excite \cite{hu2018squeeze} (SqEx) skip connection.
Finally, we adapt the ExFuse \cite{zhang2018exfuse} skip connection for the UNet architecture, using the decoder features as the high-level feature map fed into the semantic embedding branch, and following it up with a $9 \times 9$ global convolution.
Notably, compared to all other skip connections concatenating the encoder and decoder feature maps, ExFuse performs a residual skip connection by adding them.
Illustrations describing each skip connection used in our experiments can be found in Figure~\ref{fig:skips}.

Tables~\ref{tab:depth_metrics} and \ref{tab:secondary_metrics} present the performance of each skip connection on the M3D test set for the direct depth estimation metrics, as well as the boundary and smoothness preservation ones respectively.
Evidently, the hybrid skip connection (learnable blending factors, $K = 9$) outperforms the other skip connections approaches for dense regression in two aspects.
First, it offers the largest gain in terms of improving direct depth estimation performance.
Second, it additionally offers the more balanced performance increase compared to a vanilla UNet \cite{ronneberger2015u} across the secondary -- competing -- performance traits.
Finally, it does so at a reduced extra parameter cost (last column in Table~\ref{tab:secondary_metrics}).
While SqEx (Residual) offers an important performance boost for preserving the depth map's smoothness (boundaries), it does so at the expense of preserving boundaries (smoothness).
The (second) better balanced approach is that of ExFuse which manages to offer reasonable performance gains across all performance traits.
These comparisons can be more easily discerned in Figure~\ref{fig:spider} that illustrates radar plots across different normalized accuracy ($a$) and error ($e$) indicators for all performance axes:
\begin{align*}
\displaystyle
    d_a &= (0.2 \times (\delta_{1.05} + \delta_{1.1} + \delta_{1.25} + \delta_{1.25^2} + \delta_{1.25^3}))^{-1}\\
    b_a &= (\nicefrac{F_1^{1.0} + F_1^{0.5} + F_1^{0.25}}{3})^{-1}, s_a = (\nicefrac{\alpha_{11.25^o} + \alpha_{22.5^o} + \alpha_{30^o}}{3})^{-1}\\
    d_e &= (RMSE \times RMSLE)^{-1}, \, s_e = (RMSE^o)^{-1}\\
    b_e &= (\text{dbe}^{acc} \times \text{dbe}^{comp})^{-1}.
\end{align*}

\begin{figure}[!htbp]
\centering
\captionsetup[subfigure]{position=top,labelformat=empty}

\subfloat[]{\rotatebox[]{90}{Color~~~~~~~}\hspace{5pt}} 
\subfloat[]{\includegraphics[width=0.475\linewidth]{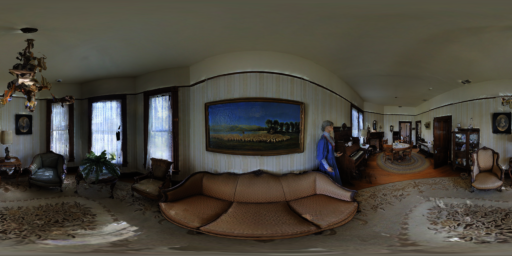}}
\subfloat[]{\includegraphics[width=0.475\linewidth]{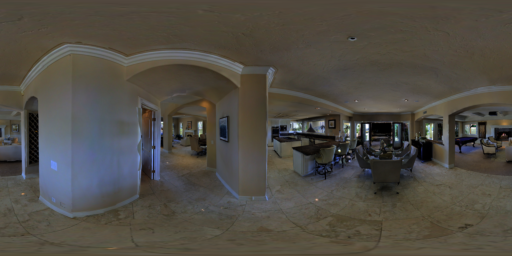}}\\
[-5.ex]
\subfloat[]{\rotatebox[]{90}{\textcolor{residual}{Residual~~~~~}}\hspace{5pt}}
\subfloat[]{\includegraphics[width=0.475\linewidth]{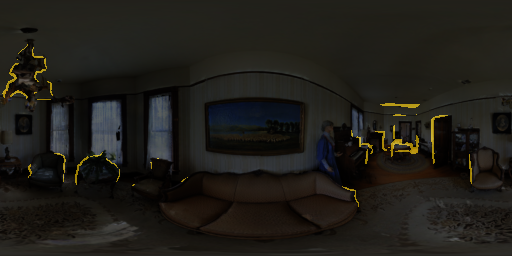}}
\subfloat[]{\includegraphics[width=0.475\linewidth]{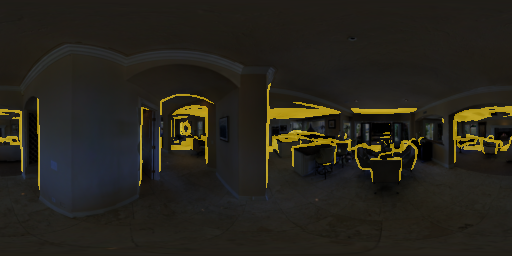}}\\
[-5.ex]
\subfloat[]{\rotatebox[]{90}{\textcolor{hybrid}{Hybrid~~~~~~}}\hspace{3pt}}
\subfloat[]{\includegraphics[width=0.475\linewidth]{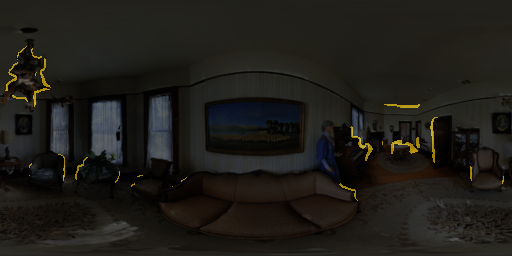}}
\subfloat[]{\includegraphics[width=0.475\linewidth]{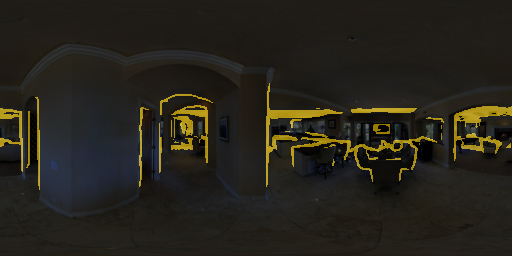}}\\
[-5.ex]
\subfloat[]{\rotatebox[]{90}{\textcolor{sqex}{SqEx~~~~~~~}}\hspace{3pt}}
\subfloat[]{\includegraphics[width=0.475\linewidth]{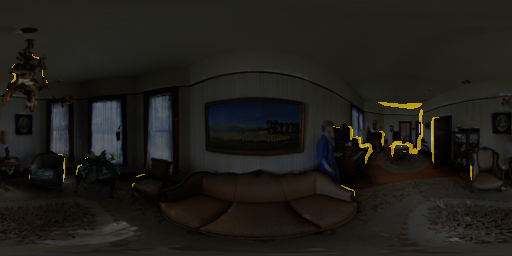}}
\subfloat[]{\includegraphics[width=0.475\linewidth]{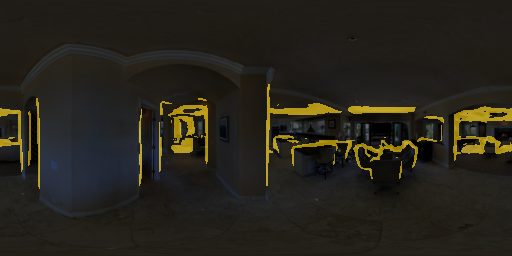}}\\
[-5.ex]
\subfloat[]{\rotatebox[]{90}{\textcolor{residual}{Residual~~~~~}}\hspace{5pt}}
\subfloat[]{\includegraphics[width=0.475\linewidth]{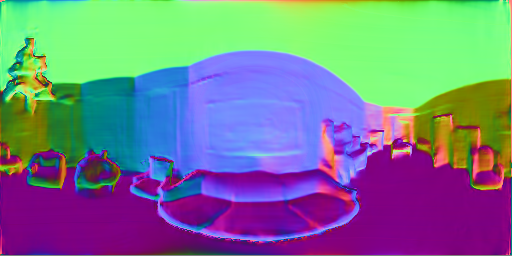}}
\subfloat[]{\includegraphics[width=0.475\linewidth]{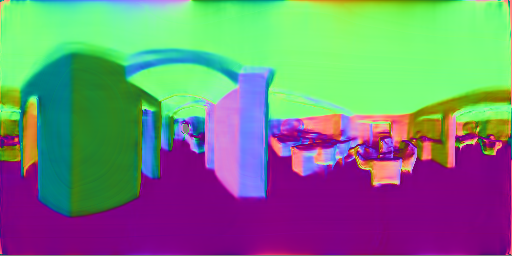}}\\
[-5.ex]
\subfloat[]{\rotatebox[]{90}{\textcolor{hybrid}{Hybrid~~~~~~}}\hspace{3pt}}
\subfloat[]{\includegraphics[width=0.475\linewidth]{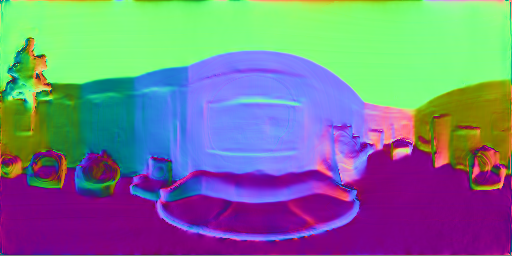}}
\subfloat[]{\includegraphics[width=0.475\linewidth]{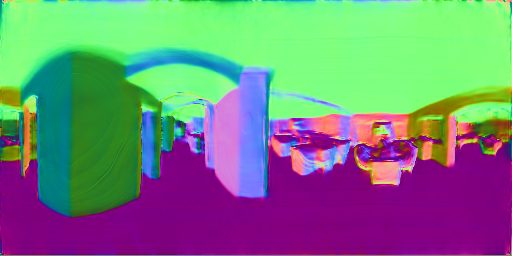}}\\
[-5.ex]
\subfloat[]{\rotatebox[]{90}{\textcolor{sqex}{SqEx~~~~~~~}}\hspace{3pt}}
\subfloat[]{\includegraphics[width=0.475\linewidth]{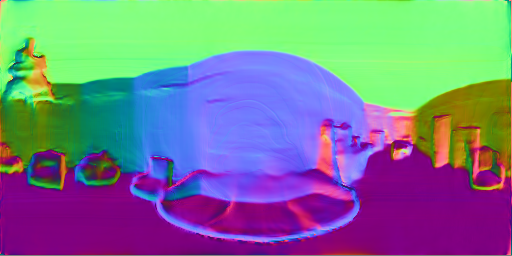}}
\subfloat[]{\includegraphics[width=0.475\linewidth]{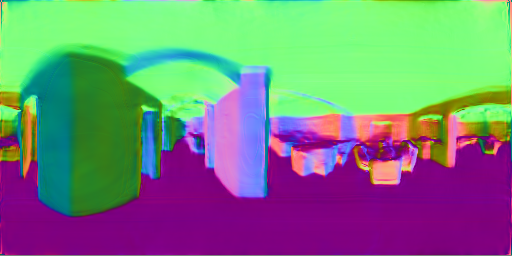}}\\
[-5.ex]
\subfloat[]{\rotatebox[]{90}{\textcolor{sqex}{Groundtruth~~}}\hspace{3pt}}
\subfloat[]{\includegraphics[width=0.475\linewidth]{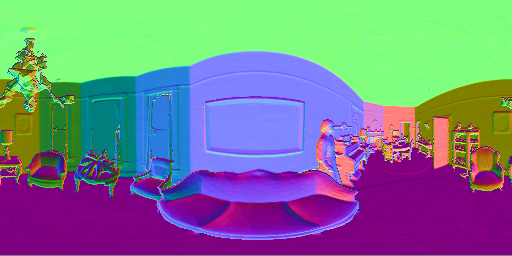}}
\subfloat[]{\includegraphics[width=0.475\linewidth]{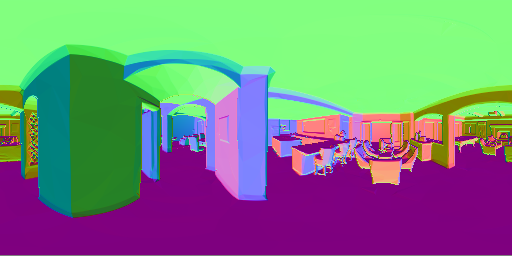}}\\
\caption{
Qualitative results on the M3D test set samples. 
From top to bottom: \textbf{i)} input color image, \textbf{ii}-\textbf{iv)} predicted depth boundaries for the \textcolor{residual}{Residual}, \textcolor{hybrid}{Hybrid} and \textcolor{sqex}{SqEx} models, \textbf{v}-\textbf{vii)} predicted normal maps for the same models, and \textbf{viii)} the groundtruth normal maps.
}

\label{fig:combo_qualitative}
\end{figure}

\noindent Figure~\ref{fig:combo_qualitative} presents qualitative results of our $K = 9$ hybrid skip model, compared to the SqEx and Residual models.
The latter are the better performing models in terms of surface and boundary preservation respectively, but clearly showcase the difficulty in achieving a balance between these two traits, as their improved performance on one, translates to a reduced performance on the other.
In contrast, the hybrid skip model strikes a better balance in preserving both traits.

\subsection{Hybrid Skip Ablation}
\label{subsec:ablation2}
We additionally perform an ablation study of the three functional components that jointly formulate the HybridSkip connection. 
First, we examine a scenario where only the learnable blending of the encoder and decoder features is introduced, denoted as $\mathcal{F}_{blend} = H_i([\boldsymbol{\delta} \, \mathcal{D} + (\mathbf{1} - \boldsymbol{\delta}) \mathcal{E}; \boldsymbol{\epsilon} \, \mathcal{E} + (\mathbf{1} - \boldsymbol{\epsilon})\mathcal{D}])$.
Then we also conduct two experiments where only a single filter is applied either only at the encoder features (low pass) or the decoder ones (high pass) respectively denoted as $\mathcal{F}_{low} = H_i([\mathbf{f}_{l}(\mathcal{E}); \mathcal{D}])$ and $\mathcal{F}_{high} = H_i([\mathcal{E}; \mathbf{f}_{h}(\mathcal{D})])$.
The results for the two larger kernels (\textit{i.e.}~$K=7, K=9$) are presented in Tables \ref{tab:ablation_direct} and \ref{tab:albation_secondary}, where the former includes the metrics related to direct depth estimation performance and the latter includes the metrics related to the secondary traits, depth smoothness and boundary preservation.

\begin{table*}[!htbp]
\centering
\caption{Direct depth metrics performance metrics for the HybridSkip connection ablation experiments. Same colorization scheme as Table \ref{tab:depth_metrics}. Since $\mathcal{F}_{blend}$ is a (learnable) blending of the encoder and decoder features, with no spatial filters applied, we duplicate the row and adjust the colorized ranking only with respect to the two different kernel sizes.}
\label{tab:ablation_direct}
\begin{tabular}{l|c|ccccccccc}
\hline
\multicolumn{1}{c|}{\multirow{3}{*}{Model}} &
  \multirow{3}{*}{Kernel} &
  \multicolumn{9}{c}{Direct Depth} \\
\multicolumn{1}{c|}{} &
   &
  \multicolumn{4}{c}{Error $\downarrow$} &
  \multicolumn{5}{c}{Accuracy $\uparrow$} \\
\multicolumn{1}{c|}{} &
   &
  \textit{RMSE} &
  \textit{RMSLE} &
  \textit{AbsRel} &
  \multicolumn{1}{c|}{\textit{SqRel}} &
  $\delta_{1.25}$ &
  $\delta_{1.25^2}$ &
  $\delta_{1.25^3}$ &
  $\delta_{1.05}$ &
  $\delta_{1.1}$ \\ \hline
$\mathcal{F}_{hybrid}$ &
  \multirow{3}{*}{$K=9$} &
  \second{0.3937} &
  \first{0.0639} &
  \first{0.1010} &
  \first{0.0596} &
  \first{90.76\%} &
  \first{97.72\%} &
  \first{99.17\%} &
  \first{38.75\%} &
  \first{64.41\%} \\
$\mathcal{F}_{low}$ &
   &
  \first{0.3921} &
  0.0718 &
  \third{0.1095} &
  \third{0.0658} &
  89.24\% &
  \third{97.50\%} &
  \second{99.14\%} &
  \second{38.47\%} &
  \second{62.58\%} \\
$\mathcal{F}_{high}$ &
   &
  0.4105 &
  \third{0.0678} &
  \second{0.1090} &
  0.0660 &
  \second{89.36\%} &
  97.43\% &
  99.07\% &
  34.77\% &
  \third{61.38\%} \\ \cline{1-2}
  \multirow{2}{*}{$\mathcal{F}_{blend}$} &
  \multirow{2}{*}{N / A} &
  \third{0.4006} &
  \second{0.0670} &
  \third{0.1095} &
  \second{0.0649} &
  \third{89.35\%} &
  \second{97.54\%} &
  \third{99.12\%} &
  \third{35.78\%} &
  61.09\% \\ \cline{3-11}
 &
   &
  \third{0.4006} &
  \second{0.0670} &
  \second{0.1095} &
  \second{0.0649} &
  \second{89.35\%} &
  \second{97.54\%} &
  \second{99.12\%} &
  35.78\% &
  61.09\% \\ \cline{1-2}
$\mathcal{F}_{hybrid}$ &
  \multirow{3}{*}{$K=7$} &
  \first{0.3912} &
  \first{0.0646} &
  \first{0.1039} &
  \first{0.0611} &
  \first{90.40\%} &
  \first{97.69\%} &
  \first{99.16\%} &
  \third{36.46\%} &
  \second{62.66\%} \\
$\mathcal{F}_{low}$ &
   &
  0.4017 &
  \third{0.0684} &
  \third{0.1106} &
  0.0698 &
  88.86\% &
  97.22\% &
  98.96\% &
  \first{39.57\%} &
  \first{62.76\%} \\
$\mathcal{F}_{high}$ &
   &
  \second{0.4002} &
  0.1200 &
  \third{0.1106} &
  \third{0.0681} &
  \third{89.06\%} &
  \third{97.36\%} &
  \third{99.05\%} &
  \second{36.49\%} &
  \third{61.43\%} \\ \hline
\end{tabular}
\end{table*}
\begin{table*}[!htbp]
\centering
\caption{Depth boundary and smoothness preservation metrics for the HybridSkip connection ablation experiments. Same colorization and arrangement scheme as Table \ref{tab:depth_metrics}.}
\label{tab:albation_secondary}
\begin{tabular}{l|c|ccccccccc}
\hline
\multicolumn{1}{c|}{\multirow{3}{*}{Model}} &
  \multicolumn{1}{l|}{\multirow{3}{*}{Kernel}} &
  \multicolumn{5}{c}{Depth Discontinuity} &
  \multicolumn{4}{c}{Depth Smoothness} \\
\multicolumn{1}{c|}{} &
  \multicolumn{1}{l|}{} &
  \multicolumn{2}{c}{Error $\downarrow$} &
  \multicolumn{3}{c}{Accuracy $\uparrow$} &
  \multicolumn{3}{c}{Accuracy $\uparrow$} &
  Error $\downarrow$ \\
\multicolumn{1}{c|}{} &
  \multicolumn{1}{l|}{} &
  \textit{dbe}\textsuperscript{acc} &
  \multicolumn{1}{c|}{\textit{dbe}\textsuperscript{comp}} &
  $F_{1}^{0.25}$ &
  $F_{1}^{0.5}$ &
  \multicolumn{1}{c|}{$F_{1}^{1}$} &
  $\alpha_{11.25^o}$ &
  $\alpha_{22.5^o}$ &
  \multicolumn{1}{c|}{$\alpha_{30^o}$} &
  \textit{RMSE\textsuperscript{o}} \\ \hline
$\mathcal{F}_{hybrid}$ &
  \multirow{3}{*}{$K=9$} &
  \second{1.312} &
  \first{3.733} &
  \third{49.41\%} &
  \third{42.94\%} &
  \multicolumn{1}{c|}{\third{34.42\%}} &
  \first{64.24\%} &
  \first{78.82\%} &
  \first{83.86\%} &
  \first{15.36} \\
$\mathcal{F}_{low}$ &
   &
  \third{1.360} &
  \third{3.960} &
  \second{50.26\%} &
  \second{43.60\%} &
  \multicolumn{1}{c|}{\second{35.62\%}} &
  \second{63.76\%} &
  \second{78.43\%} &
  \second{83.52\%} &
  \second{15.65} \\
$\mathcal{F}_{high}$ &
   &
  1.371 &
  \second{3.833} &
  47.86\% &
  41.07\% &
  \multicolumn{1}{c|}{30.56\%} &
  63.05\% &
  77.90\% &
  83.04\% &
  15.93 \\ \cline{1-2}
\multirow{2}{*}{$\mathcal{F}_{blend}$} &
   \multirow{2}{*}{N / A} &
  \first{1.308} &
  4.098 &
  \first{50.74\%} &
  \first{44.66\%} &
  \multicolumn{1}{c|}{\first{36.25\%}} &
  \third{63.09\%} &
  \third{78.04\%} &
  \third{83.23\%} &
  \third{15.90} \\ \cline{3-11}
 &
   &
  \second{1.308} &
  \third{4.098} &
  50.74\% &
  \second{44.66\%} &
  \multicolumn{1}{c|}{\second{36.25\%}} &
  \second{63.09\%} &
  \second{78.04\%} &
  \second{83.23\%} &
  \second{15.90} \\ \cline{1-2}
$\mathcal{F}_{hybrid}$ &
  \multirow{3}{*}{$K=7$} &
  1.661 &
  4.472 &
  \third{51.10\%} &
  44.14\% &
  \multicolumn{1}{c|}{35.95\%} &
  \first{63.97\%} &
  \first{78.72\%} &
  \first{83.79\%} &
  \first{15.45} \\
$\mathcal{F}_{low}$ &
   &
  \third{1.377} &
  \first{3.786} &
  \second{51.36\%} &
  \third{44.54\%} &
  \multicolumn{1}{c|}{\third{35.97\%}} &
  \third{62.99\%} &
  \third{77.80\%} &
  \third{82.97\%} &
  \third{16.04} \\
$\mathcal{F}_{high}$ &
   &
  \first{1.266} &
  \second{3.941} &
  \first{52.14\%} &
  \first{45.94\%} &
  \multicolumn{1}{c|}{\first{37.78\%}} &
  61.99\% &
  77.42\% &
  82.72\% &
  16.33 \\ \hline
\end{tabular}
\end{table*}

While each functional component in isolation may improve performance along a single axis or metric, it is evident that their combination leads to the most balanced performance boost.
Interestingly, we observe that the preservation of structural edges is easy to achieve, but at the expense of smoothness or direct performance, something that is better mitigated when all components co-exist as a HybridSkip connection.
However, the discrepancy with respect to boundary preservation between $K=7$ and $K=9$, with the smaller kernel showing improved accuracy, indicates the selection of the kernel parameters should be tuned on a per-dataset basis.

\begin{table*}[!htbp]
\centering
\caption{Direct depth metrics performance across different architectures. Same colorization scheme as Table~\ref{tab:depth_metrics}.}
\label{tab:sota_depth_metrics}
\resizebox{\linewidth}{!}{%
\begin{tabular}{l|ccccccccc}
\hline
\multicolumn{1}{c|}{\multirow{3}{*}{Model}} &
  \multicolumn{9}{c}{Direct Depth} \\
\multicolumn{1}{c|}{} &
  \multicolumn{4}{c}{Error $\downarrow$} &
  \multicolumn{5}{c}{Accuracy $\uparrow$} \\
\multicolumn{1}{c|}{} &
  \textit{RMSE} &
  \textit{RMSLE} &
  \textit{AbsRel} &
  \multicolumn{1}{c|}{\textit{SqRel}} &
  $\delta_{1.25}$ &
  $\delta_{1.25^2}$ &
  $\delta_{1.25^3}$ &
  $\delta_{1.05}$ &
  $\delta_{1.1}$ \\ \hline
\textcolor{vanilla}{Vanilla UNet} \cite{ronneberger2015u} &
  \third{0.4055} &
  0.1158 &
  \third{0.1083} &
  \second{0.0649} &
  89.43\% &
  97.34\% &
  \second{99.09\%} &
  36.67\% &
  62.12\% \\
\textcolor{bifuse}{BiFuse \cite{wang2020bifuse}} &
  0.4243 &
  \third{0.0668} &
  0.1142 &
  0.1427 &
  \third{90.68\%} &
  \third{97.22\%} &
  98.71\% &
  \second{41.18\%}	&
  \second{65.36\%} \\
\textcolor{hohonet}{HoHoNet \cite{sun2021hohonet}} &
  \first{0.3718} &
  \first{0.0603} &
  \first{0.0998} &
  \third{0.0871} &
  \first{92.14\%} &
  \first{97.80\%} &
  \third{99.08\%} &
  \first{42.65\%} &
  \first{69.72\%} \\
\textcolor{unetpp}{UNet++ \cite{10.1007/978-3-030-00889-5_1}} &
  0.4544 &
  0.0736 &
  0.1236 &
  0.1274 &
  87.81\% &
  96.65\% &
  98.49\% &
  37.18\% &
  60.94\% \\
\textcolor{attention-unetpp}{SqEx UNet++ \cite{10.1007/978-3-030-00889-5_1,hu2018squeeze}} &
  0.4507 &
  0.0716 &
  0.1231 &
  0.1526 &
  88.76\% &
  96.80\% &
  98.57\% &
  37.20\% &
  62.87\% \\
\textcolor{hybrid}{HybridSkip UNet (Ours)} &
  \second{0.3937} &
  \second{0.0639} &
  \second{0.1010} &
  \first{0.0596} &
  \second{90.76\%} &
  \second{97.72\%} &
  \first{99.17\%} &
  \third{38.75\%} &
  \third{64.41\%} \\ \hline
\end{tabular}%
}
\end{table*}
\begin{table*}[!htbp]
\centering
\caption{Number of parameters, depth boundary and smoothness preservation metrics for different architectures. Same colorization scheme as Table~\ref{tab:depth_metrics}.}
\label{tab:sota_secondary_metrics}
\resizebox{\linewidth}{!}{%
\begin{tabular}{lccccccccc}
\hline
\multicolumn{1}{c}{\multirow{3}{*}{Model}} & \multicolumn{5}{c}{Depth Discontinuity} & \multicolumn{4}{c}{Depth Smoothness} 
\\ \cline{2-10} 
\multicolumn{1}{c}{} & \multicolumn{2}{c}{Error $\downarrow$} & \multicolumn{3}{c}{Accuracy $\uparrow$} & \multicolumn{3}{c}{Accuracy $\uparrow$} & Error $\downarrow$ 
\\
\multicolumn{1}{c}{} & \textit{dbe}\textsuperscript{acc} & \multicolumn{1}{c|}{\textit{dbe}\textsuperscript{comp}} & $F_{1}^{0.25}$ & $F_{1}^{0.5}$ & \multicolumn{1}{c|}{$F_{1}^{1}$} & $\alpha_{11.25^o}$ & $\alpha_{22.5^o}$ & \multicolumn{1}{c|}{$\alpha_{30^o}$} & 
\multicolumn{1}{c}{\textit{RMSE\textsuperscript{o}}} 
\\ \hline
\multicolumn{1}{l|}{\textcolor{vanilla}{Vanilla UNet \cite{ronneberger2015u}}} &
1.279 &
4.110 &
\second{48.89\%} &
42.14\% &
\multicolumn{1}{c|}{32.33\%} &
63.02\% &
77.94\% & 
83.12\% &
\multicolumn{1}{c}{15.95}
\\
\multicolumn{1}{l|}{\textcolor{bifuse}{BiFuse \cite{wang2020bifuse}}} &
1.321 &
\first{3.580} &
41.42\% &
33.83\% &
\multicolumn{1}{c|}{28.70\%} &
\first{69.73\%} &
\first{80.98\%} &
\first{84.99\%}	&
\multicolumn{1}{c}{\first{13.89}} 
\\
\multicolumn{1}{l|}{\textcolor{hohonet}{HoHoNet \cite{sun2021hohonet}}} &
\first{1.109} &
4.019 &
45.10\% &
36.33\% &
\multicolumn{1}{c|}{30.50\%} &
55.63\% &
72.86\% &
79.10\% &
\multicolumn{1}{c}{18.75}
\\
\multicolumn{1}{l|}{\textcolor{unetpp}{UNet++ \cite{10.1007/978-3-030-00889-5_1}}} &
\third{1.235} &
3.986 & 
\third{48.18\%} &
\second{42.61\%} & 
\multicolumn{1}{c|}{\third{34.27\%}} &
62.98\% &
77.20\% &
82.37\% &
\multicolumn{1}{c}{16.27}
\\
\multicolumn{1}{l|}{\textcolor{attention-unetpp}{SqEx UNet++ \cite{10.1007/978-3-030-00889-5_1,hu2018squeeze}}} &
\second{1.144} &
\third{3.982} &
48.00\% &
\third{42.49\%} &
\multicolumn{1}{c|}{\first{34.82\%}} &
\second{66.14\%} &
\second{79.15\%} &
\third{83.85\%} &
\multicolumn{1}{c}{\second{14.98}}
\\
\multicolumn{1}{l|}{\textcolor{hybrid}{HybridSkip UNet (Ours)}} &
1.312 &
\second{3.733} &
\first{49.41\%} &
\first{42.94\%} &
\multicolumn{1}{c|}{\second{34.42\%}} &
\third{64.24\%} &
\third{78.82\%} &
\second{83.86\%} &
\multicolumn{1}{c}{\third{15.36}}
\\ \hline
\end{tabular}
}
\end{table*}
\begin{figure*}[!htbp]
\captionsetup[subfigure]{justification=centering}
\centering

\subfloat[\textcolor{hybrid}{Hybrid} \\vs\\ \textcolor{vanilla}{UNet \cite{ronneberger2015u}}\\ (\textcolor{hybrid}{0.719} vs \textcolor{vanilla}{0.206})]
{\includegraphics[width=0.195\linewidth]{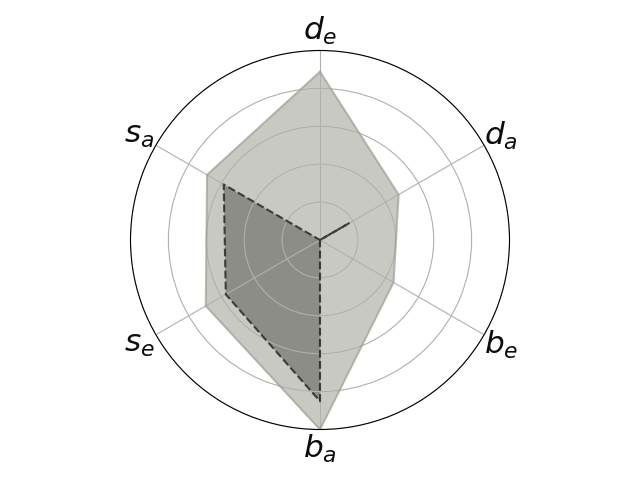}}
\subfloat[\textcolor{hybrid}{Hybrid} \\vs\\ \textcolor{bifuse}{BiFuse \cite{wang2020bifuse} }\\(\textcolor{hybrid}{0.719} vs \textcolor{bifuse}{0.651})]
{\includegraphics[width=0.195\linewidth]{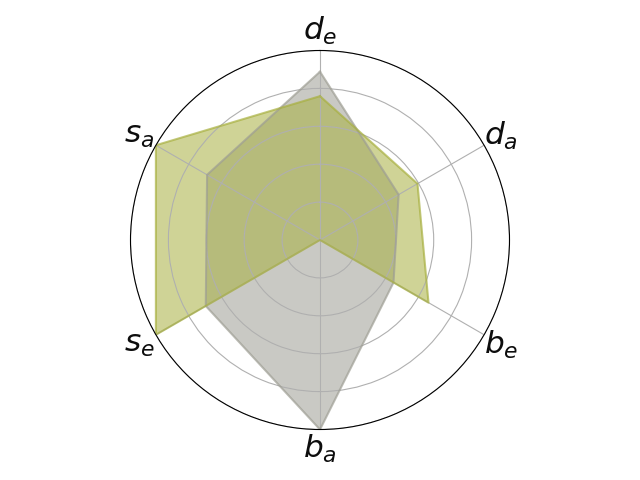}}
\subfloat[\textcolor{hybrid}{Hybrid} \\vs\\ \textcolor{hohonet}{HoHoNet \cite{sun2021hohonet}}\\(\textcolor{hybrid}{0.719} vs \textcolor{hohonet}{0.587})]
{\includegraphics[width=0.195\linewidth]{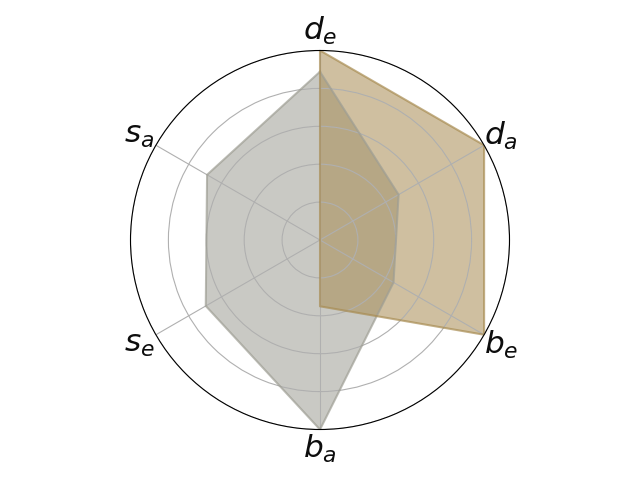}}
\subfloat[\textcolor{hybrid}{Hybrid} \\vs\\ \textcolor{unetpp}{UNet++ \cite{10.1007/978-3-030-00889-5_1}}\\(\textcolor{hybrid}{0.719} vs \textcolor{unetpp}{0.355})]
{\includegraphics[width=0.195\linewidth]{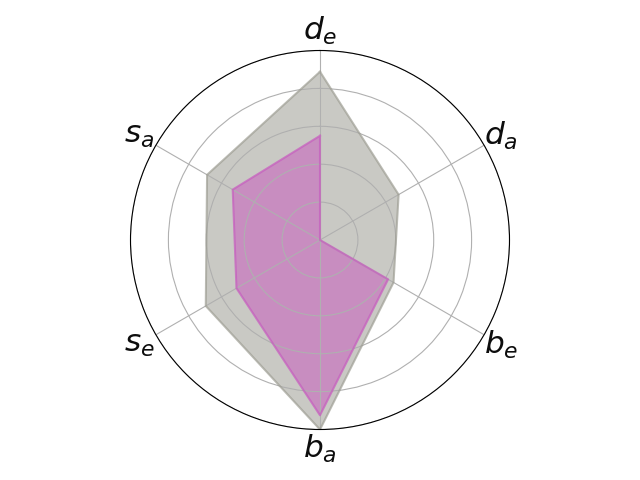}}
\subfloat[\textcolor{hybrid}{Hybrid} \\vs\\ \textcolor{attention-unetpp}{SqEx UNet++ \cite{10.1007/978-3-030-00889-5_1,hu2018squeeze}}\\(\textcolor{hybrid}{0.719} vs \textcolor{attention-unetpp}{0.706})]
{\includegraphics[width=0.195\linewidth]{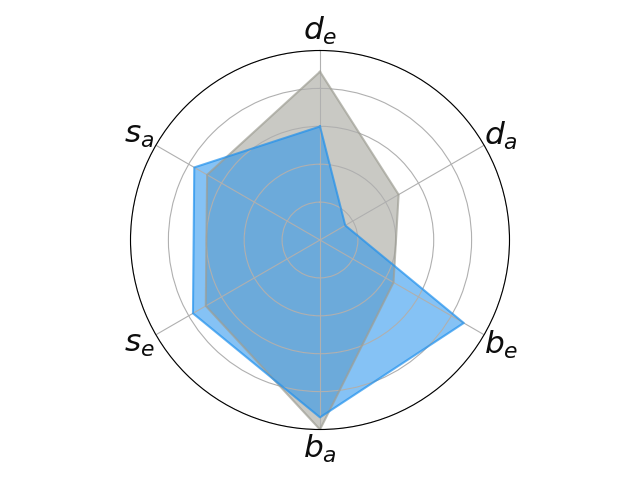}}

\vspace{10pt}

\caption{
Same scheme as Figure~\ref{fig:spider}, with larger numbers inside the  parenthesis indicating the area covered by each different approach, with larger areas indicating more balanced performance across all traits.
}
\label{fig:sota_spider}
\end{figure*}

\subsection{Other Architectures}
\label{subsec:architectures}
Finally, we examine the behavior of other UNet architectures, and established models for $360^o$ depth estimation with respect to their preservation of additional estimated signal traits.
Specifically, for the former, we use UNet++ \cite{10.1007/978-3-030-00889-5_1} and a SqEx UNet++, which is a UNet++ extended with squeeze-n-excite \cite{hu2018squeeze} skip connections, which were found to be the most balanced alternative in the skip comparison experiments in Section \ref{subsec:comparison}.
For the latter, we employ the state-of-the-art BiFuse \cite{wang2020bifuse} and HoHoNet \cite{sun2021hohonet} models.
All experiments are done using the same training scheme and our rich supervision, as presented in the previous experiments, essentially only switching the architecture for each different experiment, even for the BiFuse and HoHoNet models, for a fairer comparison.

Tables~\ref{tab:sota_depth_metrics} and \ref{tab:sota_secondary_metrics} present the direct and secondary metrics respectively, including the baseline UNet and our proposed vanilla UNet variant with HybridSkip connections.
While HoHoNet, a model specialized for the $360^o$ domain, produces high quality depth estimation, followed by our model, its behaviour with respect to preserving discontinuity and smoothness is largely reduced, showcasing worse performance even compared to the vanilla UNet.
On the other hand, BiFuse largely favours smoothness instead of boundary preservation, whereas both UNet++ variants naturally show improved boundary preservation.
As also seen in the experiments comparison different skip connections, the SqEx UNet++ balances the two secondary traits better, offering good results for smoothness as well, compared to the pure UNet++ architecture, overcoming the deficits of skip connections.
Nonetheless, its direct depth estimation performance is still at similar levels to UNet++, and inferior to the better performing $360^o$ depth estimation models.
Overall, our hybrid skip connection vanilla UNet model, offers the more balanced performance across direct and secondary trait metrics, as illustrated in Figure~\ref{fig:sota_spider}, with only the SqEx UNet++ model coming close in terms of balanced performance.

\section{Conclusion}
In this work we have designed a hybrid skip connection for the UNet architecture which relies on long range skip connections fusing features with a large semantic and spectral gap.
The simultaneous blending and spatial nature of the hybrid skip connection allows for a balanced performance boost across all performance traits for depth estimation, a dense regression task, with minimal parameter overhead.
These results indicate that it may be worth exploring the hybrid image concept in the various existing UNet modifications \cite{10.1007/978-3-030-00889-5_1,QIN2020107404,9053405,milletari2016v,valanarasu2020kiu}, or even CNN architectures without long range skip connections, with a recent report \cite{borji2022cnns} providing interesting evidence about their interplay with CNNs.
Potential explorations may include short range skip connections (e.g. residual units), or integration within basic CNN buildings blocks (e.g. squeeze-and-excite operations or Octave Convolutions \cite{chen2019drop}).
One limitation is the design of the filters themselves, which are currently performed on the spatial domain and whose parameters remain fixed during training.
While larger kernel sizes may provide a more balanced performance improvement as illustrated in Figure~\ref{fig:ablation}, each output signal's distribution may be more tuned to specific kernel parameters (\textit{e.g.}~$K = 7$ showing better boundary preservation in Table~\ref{tab:albation_secondary}).
Spectral or learnable filtering may allow models to better adapt to the task and data at hand.
Further, experimenting with adaptive blending will open up dynamic dual feature representations instead of the fixed blending factors that statically choose representations at the end of the model's training.
It also remains to be seen if these balanced skip connections also boost performance in other downstream tasks like segmentation.

\section*{Acknowledgements}
\noindent This work was supported by the European Commission H2020 funded project ATLANTIS (\href{http://atlantis-ar.eu/}{http://atlantis-ar.eu/}) [GA 951900].

{\small
\bibliographystyle{ieee_fullname}
\bibliography{egbib}
}

\end{document}